\documentclass{article}

 \usepackage[preprint]{neurips_2026}

\usepackage[utf8]{inputenc} % allow utf-8 input
\usepackage[T1]{fontenc}    % use 8-bit T1 fonts
\usepackage{hyperref}       % hyperlinks
\usepackage{url}            % simple URL typesetting
\usepackage{booktabs}       % professional-quality tables
\usepackage{amsfonts}       % blackboard math symbols
\usepackage{nicefrac}       % compact symbols for 1/2, etc.
\usepackage{microtype}      % microtypography
\usepackage{xcolor}         % colors
\usepackage{graphicx} 
\usepackage{multirow} 
\usepackage{amsmath}
\usepackage{colortbl}
\usepackage{makecell}
\usepackage{caption}
\usepackage{orcidlink}
\usepackage{diagbox}
\usepackage{subcaption}
\usepackage{pifont}
\usepackage{caption}
\newcommand{\xmark}{\ding{55}}
\usepackage[table]{xcolor}
\usepackage[normalem]{ulem}
\usepackage[accsupp]{axessibility}
\usepackage{xcolor}        
\usepackage{colortbl}     
\usepackage{array}
\usepackage{multirow}
\usepackage{caption}       
\usepackage{booktabs}     

\newcounter{blfootnote}
\newcommand\blfootnote[1]{%
  \begingroup
  \stepcounter{blfootnote}%
  \renewcommand{\thefootnote}{}%
  \renewcommand{\theHfootnote}{blfootnote.\arabic{blfootnote}}%
  \footnotetext{#1}%
  \endgroup
}
\title{HDRFace: Rethinking Face Restoration with High-Dimensional Representation}

\author{%
  Zirui Wang\textsuperscript{\rm 1},  Xianhui Lin\textsuperscript{\rm 2}$^{\dagger}$, Yi Dong\textsuperscript{\rm 2},  Bo Wei\textsuperscript{\rm 2}, Gangjian Zhang\textsuperscript{\rm 2},\\  
  \textbf{Siteng Ma\textsuperscript{\rm 2}, \textbf{Zebiao Zheng}\textsuperscript{\rm 2}, Xing Liu\textsuperscript{\rm 2}, Hong Gu\textsuperscript{\rm 2},  Minjing Dong\textsuperscript{\rm 1}}$^*$\\
  \textsuperscript{1} City University of Hong Kong\\
  \textsuperscript{2} vivo BlueImage Lab, vivo Mobile Communication Co., Ltd\\
  {\tt\small zrwang23-c@my.cityu.edu.hk} \hspace{0.1cm}
  {\tt\small xhlin129@gmail.com} \hspace{0.1cm}
  {\tt\small minjdong@cityu.edu.hk} \hspace{0.1cm} \\
  \href{https://zirui0625.github.io/projects/HDRFace}
  {\small \textbf{\textcolor{blue}{Project Page}}} \\
}
  
\begin{document}

\maketitle

% \begin{center}
%     \includegraphics[width=\textwidth]{figures/first_image.pdf}
%     {\captionsetup{hypcap=false}
%     \captionof{figure}{Motivation and improvements of HDRFace. (Left) Existing diffusion-based methods rely solely on low-quality inputs, yielding outputs that lack fine-grained detail and faithful identity preservation. We address this by injecting high-dimensional visual representations into the generative pipeline. (Right) Our method consistently outperforms OSDFace in perceptual quality (LPIPS), identity consistency (ArcFace degree), and image quality (TOPIQ).}
%     \label{f-introduction}}
% \end{center}

\begin{abstract}

  Face restoration under complex degradations still remains an ill-posed inverse problem due to severe information loss. Although diffusion models benefit from strong generative priors, most methods still condition only on low-quality inputs, making it difficult to recover identity-critical details under heavy degradations. In this work, we propose \textit{\textbf{HDRFace}}, a \textit{\textbf{H}}igh-\textit{\textbf{D}}imensional \textit{\textbf{R}}epresentation conditioned \textit{\textbf{Face}} restoration framework that injects semantically rich priors into the conditional flow without modifying the generative backbone. Our pipeline first obtains a structurally reliable intermediate restoration with an off-the-shelf restorer, then uses a pretrained high-dimensional feature encoder to extract fine-grained facial representations from both the low-quality input and the intermediate result, and injects them as additional conditions for generation. We further introduce SDFM, a Structure-Detail aware adaptive Fusion Mechanism that emphasizes global constraints during structure modeling and strengthens representation guidance during detail synthesis, balancing structural consistency and detail fidelity. To validate the generalization ability of our method, we implement the proposed framework on two generative models, SD V2.1-base and Qwen-Image, and consistently observe stable and coherent performance gains across different architectures.
\end{abstract}

\begin{figure}
\centering
\includegraphics[width=\linewidth]{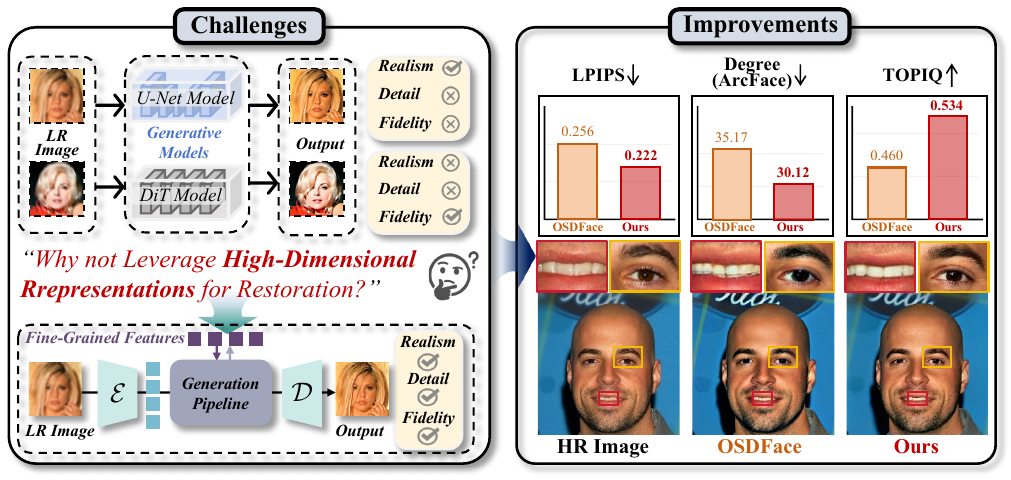}
\caption{ Motivation and improvements of HDRFace. 
(Left) Existing diffusion-based methods rely solely on low-quality inputs, yielding outputs that lack fine-grained detail and faithful identity preservation. We address this by injecting high-dimensional visual representations into the generative pipeline. (Right) Our method consistently outperforms OSDFace in perceptual quality (LPIPS), identity consistency (ArcFace degree), and image quality (TOPIQ).
}
\label{f-introduction}
\end{figure}

\blfootnote{$^{\dagger}$ Project leader. $^*$ Corresponding author.}

\section{Introduction}
Face restoration aims to reconstruct high-quality face images from low-quality inputs that have undergone complex degradations. Degradations such as blur, noise, downsampling, and JPEG compression can severely damage facial structures and fine-grained texture details, making the task inherently an ill-posed inverse problem. In recent years, diffusion models~\cite{ho2020denoising, rombach2022high}, empowered by strong generative priors and superior detail completion capabilities, have achieved notable progress in improving perceptual realism and have gradually become a dominant paradigm in face restoration~\cite{do2025dynfacerestore,yang2023pgdiff,lin2024diffbir,wang2025osdface, yue2024difface, liang2025authface}, particularly demonstrating enhanced robustness and advantages under severe degradation conditions.

However, most existing diffusion-based face restoration methods~\cite{wang2025osdface, yang2023pgdiff, yue2024difface, liang2025authface} still exhibit suboptimal performance in fine-grained detail reconstruction and image fidelity. The fundamental reason lies in the fact that these approaches typically rely solely on low-quality images as the only input and attempt to learn a direct mapping from LQ to HQ images. Due to the severe information loss and complex degradations commonly present in low-quality observations, many identity-related facial details are irreversibly damaged or heavily blurred, making it difficult for the model to accurately infer their true structures and textures from only LQ input. As a result, the reconstructed images often fail to achieve satisfactory identity consistency and perceptual realism.

To mitigate the aforementioned limitations, we seek a solution that introduces more reliable priors beyond the low-quality input to more effectively constrain the generation process. Our core objectives are twofold: how to obtain fine-grained high-dimensional visual representations, and how to inject them into the generative model to guide the reconstruction of fine-grained details. A natural approach is to first leverage an existing face restoration baseline model to obtain an intermediate restored image, which, compared to the original low-quality input, contains richer facial detail features and can provide more refined visual guidance for subsequent generation, thereby alleviating the ill-posedness caused by complex degradations. Furthermore, employing a powerful representation encoder to extract high-dimensional facial representations from this intermediate result can provide richer and more discriminative visual guidance for identity- and detail-sensitive reconstruction. However, effectively integrating such high-dimensional representations into the generation pipeline is non-trivial and requires careful consideration of several key questions. ~\textbf{First}, how can the model selectively exploit information relevant to identity and structure while suppressing redundancy and noise? ~\textbf{Second}, whether the original low-quality input should serve as a joint conditional input alongside the intermediate result, and if so, how to reconcile the feature discrepancy between the two? ~\textbf{Finally}, how can the overall framework maintain strong generalization capability without being tied to a specific generative backbone?

Based on the above analysis and discussion, we propose HDRFace, a face restoration method that introduces High-Dimensional Representation features via the conditional branch of a generative model. We first use an existing face restoration model to perform an initial restoration on low-quality face images, obtaining intermediate results with more complete structural information; then, we use a high-dimensional representation encoder DINOv3 to extract fine-grained and semantically rich high-dimensional representations from the low-quality input and the initially restored images, and inject them as conditional information into the base generative model to provide a more discriminative prior for subsequent generation. To better fuse multi-source information, we design an SDFM module, a Structure–Detail aware adaptive Fusion Mechanism that adaptively emphasizes the global constraints from the low-quality input during structure modeling and strengthens the fine-grained representation features during detail modeling, thereby balancing structural consistency and detail realism. 
We implemented and validated the proposed method on SD V2.1-base and Qwen-Image, which represent two mainstream generative paradigms: a diffusion-denoising U-Net architecture and a DiT architecture based on Rectified Flow, respectively. Experimental results demonstrate that the method yields stable and consistent improvements in restoration quality across different model architectures and generation mechanisms, highlighting strong cross-architecture generalization and practical applicability. 

To summarize, the contributions of HDRFace are as follows:
\begin{itemize}
    \item We propose a high-dimensional representation conditioned face restoration framework that injects DINOv3 semantic features into the conditional branch to provide priors beyond low-quality inputs and ease the ill-posedness caused by missing information.

    \item We design an architecture-independent module SDFM that adaptively fuses low-quality inputs with high-dimensional features, balancing structural consistency and detail fidelity without changing the generative backbone.

    \item Extensive experiments on SD V2.1-base and Qwen-Image demonstrate strong generalization and architecture independence with consistent restoration quality gains.
\end{itemize}
\section{Related work}
\subsection{Face Restoration}
Face restoration aims to recover high-quality (HQ) face images from low-quality (LQ) inputs affected by complex degradations. Such degradations often destroy key features and details in the original facial images, making the task more challenging. Traditional methods~\cite{chen2018fsrnet, shen2018deep, yu2018super, chen2021progressive, li2020blind} typically rely on geometric priors for face restoration, but they remain limited in reconstructing high-quality details. In recent years, leveraging generative priors from diffusion models for face restoration has attracted considerable attention~\cite{yang2023pgdiff,wang2025osdface,yue2024difface,liang2025authface}; thanks to their strong generative capability, diffusion models can compensate for facial details corrupted during degradation, demonstrating substantial potential for restoration.

Specifically, PGDiff~\cite{yang2023pgdiff} integrates color and structural guidance with a generative model to achieve high-quality face restoration. DifFace~\cite{yue2024difface} aligns low-quality inputs to an intermediate state of a pre-trained diffusion model and leverages the diffusion prior to achieve more robust face restoration under complex unseen degradations. OSDFace~\cite{wang2025osdface} proposes a one-step diffusion framework for face restoration, introducing prior information via a Visual Representation Embedder to perform face restoration. Although these methods achieve promising restoration performance, they still rely on LQ images as the sole input and lack fine-grained facial representations; consequently, they remain limited in detail reconstruction and consistency.

\subsection{High Fidelity Generation}
With the development of diffusion models~\cite{ho2020denoising}, latent diffusion models (LDMs)~\cite{rombach2022high} have become a dominant paradigm for generative modeling. By training diffusion models in a compact latent space learned by a VAE, LDMs substantially reduce computational cost while maintaining high generation quality. However, the compression into a low-dimensional latent space limits information capacity, which in turn constrains high-fidelity detail expression and high-quality generation. To alleviate this inherent limitation, recent works~\cite{ma2025deco, zheng2025diffusion, li2025back} have focused on improving latent representations and generative capability. In low-level vision, DINO-IR~\cite{lin2023multi} leverages DINOv2 degradation-independent features for multi-task restoration via pixel–semantic fusion, while F2IDiff~\cite{jangid2025f2idiff} uses DINOv2 features as stricter conditioning for low-hallucination, high-fidelity single image super-resolution task; however, their application to face restoration task, especially with one-step generative models, still remains underexplored.

\begin{figure}[!t]
\centering
\includegraphics[width=\linewidth]{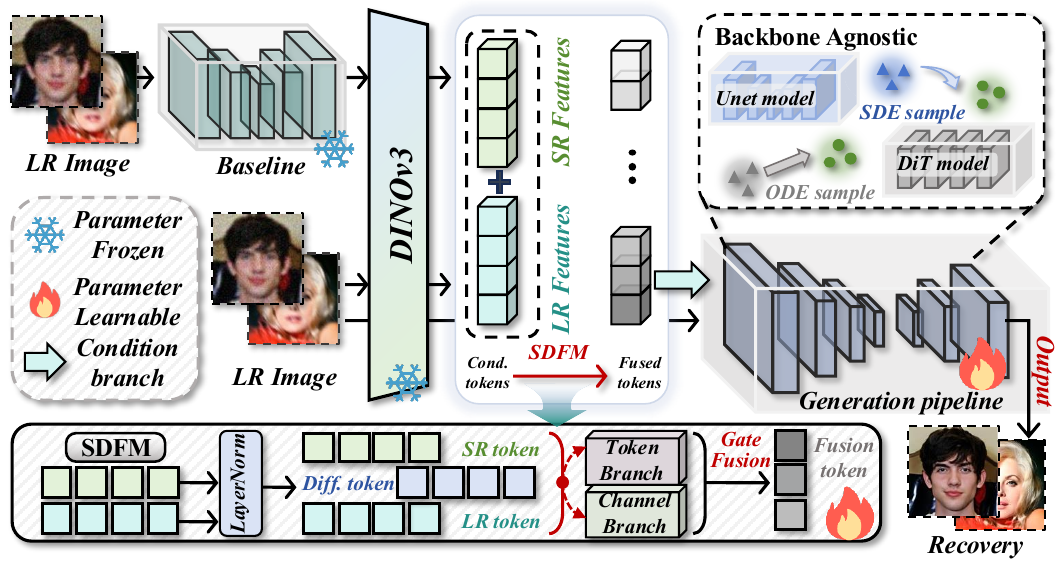}
\caption{Overview of our proposed HDRFace framework.}
\label{f-pipeline}
\vspace{-1em}
\end{figure}
\section{Methods}
In this section, we provide a systematic exposition of the proposed method by focusing on three progressively connected core questions: how to obtain high-dimensional facial representations, how to effectively incorporate such high-dimensional facial representations into the generative model, and how to train the overall restoration model. Based on these three questions, we progressively present the design and implementation details of our method. The overall framework of our approach is illustrated in Fig.~\ref{f-pipeline}.

\subsection{How to obtain High-dimensional Facial Representations}
To address this problem, we first require two key components: a high-dimensional representation encoder that can extract fine-grained facial features, and a baseline model that can provide reliable intermediate restoration results. For the representation encoder, we adopt DINOv3~\cite{simeoni2025dinov3} as the high-dimensional feature extractor. Compared with representation models such as CLIP~\cite{radford2021learning}, which emphasize image-text alignment, DINOv3 focuses more on representation learning in the visual feature space, and therefore can more effectively capture detailed facial features in images. The representation process can be formulated as follows:
\begin{equation}
\mathbf{F}_{\mathrm{face}} = \mathcal{E}_{\mathrm{DINOv3}}\!\left(\mathbf{I}_{\mathrm{mid}}\right), \quad \mathbf{F}_{\mathrm{face}} \in \mathbb{R}^{N \times d},
\end{equation}
Here, $\mathbf{I}_{\mathrm{mid}}$ denotes the intermediate restoration result, while 
$N$ and $d$ represent the number of DINO feature tokens and dimension.

% In this basis, the next key issue is how to obtain a reliable intermediate restoration result. If a pretrained model with the same architecture as the training model is used for restoration guidance, the final model’s performance is typically limited by the capacity of that pretrained model, meaning that optimization tends to make it approach the pretrained model’s performance limit rather than exceed it. As shown in Tab.~\ref{main:table}, we observe that Qwen-Image can restore facial images with high fidelity after simple training. In contrast, \cite{wang2025osdface} trained based on SD V2.1-base shows slightly lower fidelity, but produces results with richer details and better visual clarity. Owing to the complementary restoration characteristics of the two models, we use Qwen-Image as the intermediate restoration model when training the SD V2.1-base model, and conversely adopt the SD V2.1-base model when training Qwen-Image. All intermediate restoration models are trained only on LR facial images from the same dataset, without additional supervision, thus avoiding unfairness caused by data leakage.

On this basis, the next key issue is to obtain a reliable intermediate restoration result. Using a pretrained model with the same architecture as the target training model for restoration guidance may constrain the final performance, as optimization tends to approach the pretrained model’s upper bound rather than surpass it. As shown in Tab.~\ref{main:table}, Qwen-Image achieves high-fidelity facial restoration after simple training, while the SD V2.1-base model used in \cite{wang2025osdface} produces slightly lower fidelity but richer details and clearer visual results. Since the two models show complementary restoration behaviors, we use Qwen-Image as the intermediate restoration model when training SD V2.1-base, and use SD V2.1-base when training Qwen-Image. All intermediate restoration models are trained only on LR facial images from the same dataset, without additional supervision, thereby avoiding unfairness caused by data leakage.

\subsection{How to incorporate High-Dimensional Facial Representations}
% After obtaining the high-dimensional facial representations, the next step is to consider how to incorporate them as guidance into the generative model. Specifically, we treat the high-dimensional facial representations as conditional features in a form analogous to text embeddings, and feed them into the conditional branch of the generative model, which improves the generalization ability of our method. In previous methods, the conditional branch is typically conditioned on text or low-resolution images. Although high-dimensional representations can effectively replace textual guidance, we do not discard the LR image, since incorporating it helps preserve the structural fidelity of the generated results. Therefore, we also feed the LR image into DINOv3 to obtain its corresponding representation, denoted as $\mathbf{F}_{\mathrm{lr}}$:

After obtaining the high-dimensional facial representations, we further incorporate them into the generative model as guidance. Specifically, these representations are treated as conditional features, similar to text embeddings, and fed into the conditional branch to improve generalization. While previous methods usually condition this branch on text, our high-dimensional representations can effectively replace textual guidance. However, we still retain the LR image to preserve structural fidelity. Thus, we also feed the LR image into DINOv3 to extract its representation, denoted as $\mathbf{F}_{\mathrm{lr}}$.
\begin{equation}
\mathbf{F}_{\mathrm{lr}} = \mathcal{E}_{\mathrm{DINOv3}}\!\left(\mathbf{I}_{\mathrm{LR}}\right), \quad \mathbf{F}_{\mathrm{lr}} \in \mathbb{R}^{N \times d}.
\end{equation}
We expect the model to rely more on $\mathbf{F}_{\mathrm{lr}}$ for structural information, while placing greater emphasis on $\mathbf{F}_{\mathrm{face}}$ for fine-grained details. To better model the relationship between the two, we further propose the SDFM module. 

Specifically, we first normalize the two feature sequences and compute their absolute difference to explicitly characterize the discrepancy between facial-detail features and LR structural features:
\begin{equation}
\tilde{\mathbf{F}}_{\mathrm{face}} = \mathrm{LN}(\mathbf{F}_{\mathrm{face}}),\;
\tilde{\mathbf{F}}_{\mathrm{lr}} = \mathrm{LN}(\mathbf{F}_{\mathrm{lr}}),\;
\tilde{\mathbf{F}}_{\mathrm{Diff}} = \left| \tilde{\mathbf{F}}_{\mathrm{face}} - \tilde{\mathbf{F}}_{\mathrm{lr}} \right|.
\end{equation}
Here, $\mathrm{LN}$ denotes the LayerNorm operation, and $\tilde{\mathbf{F}}_{\mathrm{Diff}}$ denotes the discrepancy between the two representations. On this basis, we model the complementary relationship between $\mathbf{F}_{\mathrm{face}}$ and $\mathbf{F}_{\mathrm{lr}}$ from two perspectives, namely global channel statistics and local token interactions, so as to adaptively regulate the balance between structural information and fine-grained details. Specifically, the module first learns a channel-wise gating weight from globally pooled features:
\begin{equation}
\boldsymbol{\alpha}_{c}=\sigma\!\left(\mathrm{MLP}_{c}\!\left(\left[
\mathrm{Pool}(\tilde{\mathbf{F}}_{\mathrm{face}})\,;\,
\mathrm{Pool}(\tilde{\mathbf{F}}_{\mathrm{lr}})\,;\,
\mathrm{Pool}(\tilde{\mathbf{F}}_{\mathrm{Diff}})
\right]\right)\right),\;
\boldsymbol{\alpha}_{c} \in \mathbb{R}^{1 \times d},
\end{equation}
where $\boldsymbol{\alpha}_{c}$ denotes the channel-wise gating weight, which captures the global importance of each feature channel based on the pooled statistics of facial features, LR features, and their discrepancy. Next, the module learns a token-wise gating weight from token-level concatenated features:
\begin{equation}
\boldsymbol{\alpha}_{t}
=
\sigma\!\left(\mathrm{MLP}_{t}\!\left(\left[
\tilde{\mathbf{F}}_{\mathrm{face}}\,;\,
\tilde{\mathbf{F}}_{\mathrm{lr}}\,;\,
\tilde{\mathbf{F}}_{\mathrm{Diff}}
\right]\right)\right),
\;
\boldsymbol{\alpha}_{t} \in \mathbb{R}^{N \times 1},
\end{equation}
where $\boldsymbol{\alpha}_{t}$ is the token-wise gating weight, which characterizes the relative importance of the two representations at each token location. The final fusion weight is then obtained by combining the channel-wise and token-wise gates:
\begin{equation}
\boldsymbol{\alpha}
=
\mathrm{Clamp}\!\left(
\boldsymbol{\alpha}_{t} \odot \boldsymbol{\alpha}_{c},
\epsilon,\,1-\epsilon
\right),\;
\boldsymbol{\alpha} \in \mathbb{R}^{N \times d},
\end{equation}
where $\boldsymbol{\alpha}$ denotes the final fusion weight, and $\mathrm{Clamp}(\cdot)$ is used to constrain its value range for more stable fusion. Based on the learned fusion weight, the two representations are combined in a weighted manner to obtain the final fused feature:
\begin{equation}
\mathbf{F}_{\mathrm{fused}}
=
\boldsymbol{\alpha} \odot \tilde{\mathbf{F}}_{\mathrm{face}}
+
\left(1-\boldsymbol{\alpha}\right) \odot \tilde{\mathbf{F}}_{\mathrm{lr}}.
\end{equation}
Here, $\mathbf{F}_{\mathrm{fused}}$ denotes the fused representation, which adaptively integrates fine-grained facial details from $\tilde{\mathbf{F}}_{\mathrm{face}}$ and structural cues from $\tilde{\mathbf{F}}_{\mathrm{lr}}$. In this way, SDFM enhances the representation of fine facial details while preserving the structural consistency of the LR input.
\subsection{How to train the overall restoration model}

After obtaining the fused facial representation, we train a one-step face restoration model by using it as conditional guidance. Given an LR facial image, the model first maps it into the latent space and constructs a noisy latent input according to the corresponding generative architecture. For SD V2.1-base, the LR latent is directly treated as the noisy input at a fixed timestep, while for Qwen-Image, we follow the Rectified Flow formulation and introduce a linear noise perturbation process. Conditioned on the fused high-dimensional facial representation, the model performs one-step denoising to generate the restored high-quality face image:
\begin{equation}
\hat{\mathbf{I}}_{\mathrm{face}}
=
\mathcal{D}_{\mathrm{VAE}}
\left(
\mathcal{G}_{\theta}(\mathbf{z}_T, T, \mathbf{F}_{\mathrm{fused}})
\right),
\end{equation}
where $z_T$ denotes the noisy latent input, $T$ is the predefined timestep, $\mathbf{F}_{\mathrm{fused}}$ is the fused facial representation, and $\mathcal{D}_{\mathrm{VAE}}$ represents the VAE decoder. For Qwen-Image, we additionally introduce the noise-perturbed LR latent as an auxiliary conditional input following~\cite{fang2025one}, with implementation details provided in the Appendix~\ref{appendix:a.1}.

To optimize the restoration model, we adopt a unified training objective consisting of reconstruction, perceptual, identity, and GAN losses:
% \begin{equation}
% \mathcal{L}_{gen}
% =
% \mathcal{L}_{rec}
% +
% \lambda_{per}\mathcal{L}_{per}
% +
% \lambda_{ID}\mathcal{L}_{ID}
% +
% \lambda_{dis}\mathcal{L}_{G}.
% \end{equation}
\begin{equation}
\begin{aligned}
\mathcal{L}_{\mathrm{total}} \;=\;&
\lambda_{\mathrm{G}}\,\mathcal{L}_{G}\!\left(\mathbf{z}_{\mathrm{HR}}, \hat{\mathbf{z}}_{\mathrm{face}}\right)
+\lambda_{\mathrm{ID}}\,\mathcal{L}_{\mathrm{ID}}\!\left(\mathbf{I}_{\mathrm{HR}}, \hat{\mathbf{I}}_{\mathrm{face}}\right) \\
&+\lambda_{\mathrm{per}}\,\mathcal{L}_{\mathrm{per}}\!\left(\mathbf{I}_{\mathrm{HR}}, \hat{\mathbf{I}}_{\mathrm{face}}\right)
+\lambda_{\mathrm{rec}}\,\mathcal{L}_{\mathrm{rec}}\!\left(\mathbf{I}_{\mathrm{HR}}, \hat{\mathbf{I}}_{\mathrm{face}}\right),
\end{aligned}
\end{equation}
Here, $\mathcal{L}_\mathrm{rec}$ denotes the reconstruction loss between the restored image and the HR ground truth, $\mathcal{L}_\mathrm{per}$ improves perceptual quality, $\mathcal{L}_\mathrm{ID}$ preserves facial identity, and $\mathcal{L}_\mathrm{G}$ enhances visual realism through adversarial learning. For different generative architectures, we adopt the corresponding adversarial loss formulation, while keeping the overall optimization objective consistent. More details of the loss implementation are provided in the Appendix~\ref{appendix:a.1}.

Through this unified one-step training framework, high-dimensional facial representations are effectively incorporated into different generative architectures, enabling high-fidelity and high-quality face restoration.

\section{Experiments}
\subsection{Experimental Settings}
\textbf{Experiment datasets.} Our model is trained on the FFHQ~\cite{karras2019style} dataset, which contains 70,000 high-quality face images. All images are uniformly resized to 512×512 resolution, and we synthesize degraded samples using the same degradation pipeline as VQFR~\cite{gu2022vqfr} to generate the corresponding low-quality inputs. During evaluation, we test our method on the synthetic benchmark CelebA-Test~\cite{karras2017progressive}, which includes 3,000 different face images. The degradation generation pipeline is kept consistent with that used in training to ensure fair and comparable evaluation settings. We also conduct experiments on two real-world datasets, LFW-Test~\cite{huang2008labeled} and CelebChild~\cite{wang2021towards}, which contain diverse and complex degradations.

\noindent\textbf{Evaluation Metrics.}
We conduct experiments under multiple metrics to evaluate the performance of the proposed method. Specifically, PSNR and SSIM are used to assess pixel-level restoration fidelity, while LPIPS~\cite{zhang2018unreasonable}, DISTS~\cite{ding2020image}, and TOPIQ~\cite{chen2024topiq} are adopted as perceptual quality metrics to evaluate the fidelity of our method; and CLIP-IQA~\cite{wang2023exploring}, MUSIQ~\cite{ke2021musiq}, Q-Align~\cite{wu2023q}, and LIQE~\cite{zhang2023blind} are used as no-reference image quality assessment (IQA) metrics to evaluate the overall quality of the restored images. In addition, we use the ArcFace~\cite{deng2019arcface} embedding angle (Deg.) and landmark distance (LMD) as metrics for identity consistency, and employ FID~\cite{heusel2017gans} to measure the distribution similarity between real images and restored images.

\noindent\textbf{Implementation Details.}
Regarding the implementation details on SD V2.1-base, our method generally follows the settings of OSDFace~\cite{wang2025osdface}. Specifically, we adopt DINOv3-L~\cite{simeoni2025dinov3} as the representation extractor and fine-tune the model using LoRA~\cite{hu2022lora}, with both the rank and alpha set to 16. Training is conducted on 2 NVIDIA A100 GPUs with a batch size of 2, for a total of 100k iterations. For the training of Qwen-Image, we generally followed the training pipeline of ODTSR~\cite{fang2025one}. During training, we performed full-parameter fine-tuning with a batch size of 1, and ran 10k iterations on a single NVIDIA A100 GPU.

\noindent\textbf{Compared Methods.}
We compare HDRFace with three categories of face restoration methods, including Non-Diffusion methods such as VQFR~\cite{gu2022vqfr}, RestoreFormer++~\cite{wang2023restoreformer++}, CodeFormer~\cite{zhou2022towards}, and DAEFR~\cite{tsai2024dual}, Multi-Step Diffusion methods such as PGDiff~\cite{yang2023pgdiff}, DifFace~\cite{yue2024difface}, and AuthFace~\cite{liang2025authface}, and the One-Step Diffusion method OSDFace~\cite{wang2025osdface}.

\definecolor{retouchbg}{RGB}{245,226,204}
\definecolor{editbg}{RGB}{206,210,255}  
\definecolor{rlbg}{RGB}{244,205,214} 
\definecolor{bestcolor}{RGB}{220, 20, 60}    
\definecolor{secondcolor}{RGB}{0, 0, 255}
\newcommand{\best}[1]{\textcolor{bestcolor}{{#1}}}
\newcommand{\second}[1]{\textcolor{secondcolor}{#1}}
\begin{table*}[t]
\centering
\small
\caption{Quantitative comparison on the synthetic CelebA-Test dataset among non-diffusion, multi-step diffusion, and one-step diffusion methods. The best and second-best results are highlighted in \best{Red} and \second{Blue}.}
\label{main:table}
\setlength{\tabcolsep}{5pt}
\renewcommand{\arraystretch}{1.1}

\begin{tabular}{l|cc|ccc|ccc}
\toprule
 & \multicolumn{8}{c}{CelebA-Test}  \\
\cline{2-9}
 & SSIM$\uparrow$ & PSNR$\uparrow$ & DISTS$\downarrow$ & LPIPS$\downarrow$ & TOPIQ $\uparrow$ & Deg.$\downarrow$ & LMD$\downarrow$ & FID$\downarrow$   \\
\midrule

\rowcolor{retouchbg}\multicolumn{9}{c}{\textit{Non-Diffusion}} \\
% GFPGAN & 0.7131 & 25.01 & 4.514 & 0.1817 & 0.2604 & 0.4590 \\
VQFR~\cite{gu2022vqfr} & 0.6470 & 23.14 & 0.1643 & 0.2543 & 0.4698 & 36.80 & 2.442 & 42.99 \\
RestoreFormer++~\cite{wang2023restoreformer++} & 0.6580 & 23.67 & 0.1594 & 0.2386 & 0.5004 & \second{32.04} & 2.123 & 41.87 \\
CodeFormer~\cite{zhou2022towards} & 0.6788 & \best{24.31} & 0.1661 & 0.2307 & 0.5092 & 37.76 & 2.454 & 53.17 \\
DAEFR~\cite{tsai2024dual} & 0.5997 & 21.38 & 0.1711 & 0.2621 & 0.4242 & 46.38 & 3.059 & 42.46\\
\midrule

\rowcolor{editbg}\multicolumn{9}{c}{\textit{Multi-Step Diffusion}}\\
PGDiff~\cite{yang2023pgdiff} & 0.6555 & 22.22 & 0.1733 & 0.3046 & 0.3838 & 55.22 & 3.930 & 49.67 \\
DifFace~\cite{yue2024difface} & \second{0.6802} & 24.03 & 0.1669 & 0.2747 & 0.4294 & 44.11 & 2.777 & \best{40.11}\\
AuthFace~\cite{liang2025authface}  & 0.6550 & \second{24.31} & 0.1710 & 0.2633 & 0.4731 & 39.37 & 2.460 & 51.46 \\
\midrule

\rowcolor{rlbg}\multicolumn{9}{c}{\textit{One-Step Diffusion}}\\
OSDFace~\cite{wang2025osdface} & 0.6351 & 22.12 & 0.1660 & 0.2561 & 0.4601 & 35.71 & 2.223 & 46.34 \\
% Qwen-Image &  &  &  &  &  &  \\
Ours (SD V2.1-base)
& 0.6782 & 23.62 & \best{0.1433} & \second{0.2224} & 	\second{0.5344} & \best{30.12} & \best{1.939} & \second{40.66} \\
Ours (Qwen-Image) & \best{0.6899} & 24.08 & \second{0.1520} & \best{0.2045} & \best{0.5372} & 33.30 & \second{1.991} & 50.14 \\
\bottomrule
\end{tabular}
\end{table*}

\definecolor{retouchbg}{RGB}{245,226,204}
\definecolor{editbg}{RGB}{206,210,255}  
\definecolor{rlbg}{RGB}{244,205,214} 
\definecolor{bestcolor}{RGB}{220, 20, 60}    
\definecolor{secondcolor}{RGB}{0, 0, 255}
\begin{figure*}[t]
\centering
\begin{minipage}[t]{0.44\textwidth}
    \centering
    \captionof{table}{Comparison of metrics with our baseline methods.}
    \label{Table: compare baseline}
    \renewcommand{\arraystretch}{1.15}
    \setlength{\tabcolsep}{2pt}
    \small
    \begin{tabular}{l|cccc}
        \Xhline{1.1pt}
        \multirow{2}{*}{Methods} & \multicolumn{4}{c}{\textbf{CelebA-Test}} \\
        \cline{2-5}
        & Q-Align$\uparrow$ & LIQE$\uparrow$ & Deg.$\downarrow$ & FID$\downarrow$ \\
        \Xhline{0.8pt}
        OSDFace  & 4.452 & 4.823 & 35.71 & 46.34 \\
        % \rowcolor{retouchbg}
        Ours     & 4.544 & 4.882 & 30.12 & 40.65 \\
        \rowcolor{gray!15}
        \textit{Gain} 
                 & \textcolor{red}{+0.092} 
                 & \textcolor{red}{+0.059} 
                 & \textcolor{red}{$-$5.590} 
                 & \textcolor{red}{$-$5.690} \\
        \hline
        Qwen     & 4.391 & 4.711 & 34.56 & 50.60 \\
        % \rowcolor{editbg}
        Ours     & 4.546 & 4.841 & 33.30 & 50.14 \\
        \rowcolor{gray!15}
        \textit{Gain} 
                 & \textcolor{red}{+0.155} 
                 & \textcolor{red}{+0.130} 
                 & \textcolor{red}{$-$1.260} 
                 & \textcolor{red}{$-$0.460} \\
        \Xhline{1.1pt}
    \end{tabular}
\end{minipage}
\hfill
\begin{minipage}[t]{0.55\textwidth}
    \centering
    \captionof{table}{Quantitative comparison on real-world datasets. \best{Red} and \second{Blue} denote the best and second results.}
    \label{table:real-data}
    \renewcommand{\arraystretch}{1.15}
    \setlength{\tabcolsep}{2pt}
    \small
    \begin{tabular}{l|cc|cc}
        \Xhline{1.1pt}
        \multirow{2}{*}{Methods} 
                & \multicolumn{2}{c|}{\textbf{CelebChild}} 
            & \multicolumn{2}{c}{\textbf{LFW-Test}} \\
        \cline{2-5}
        & Q-Align$\uparrow$ & MUSIQ$\uparrow$
        & Q-Align$\uparrow$ & MUSIQ$\uparrow$ \\
        \Xhline{0.8pt}
        VQFR~\cite{gu2022vqfr}              & 3.676          & 71.72          & 4.369          & 74.75 \\
        RF++~\cite{wang2023restoreformer++} & 3.594          & 69.40          & 4.077          & 72.25 \\
        PGDiff~\cite{yang2023pgdiff}        & 3.574          & 66.73          & 4.088          & 71.47 \\
        OSDFace~\cite{wang2025osdface}      & \second{3.742} & \best{72.53}   & \second{4.374} & \second{75.35} \\
        \hline
        % \rowcolor{retouchbg}
        \rowcolor{gray!15} Ours (SD)                            & 3.584          & 68.71          & 4.236          & 73.42 \\
        % \rowcolor{editbg}
        \rowcolor{gray!15} Ours (Qwen)                          & \best{3.880}   & \second{72.46} & \best{4.435}   & \best{76.04} \\
        \Xhline{1.1pt}
    \end{tabular}
\end{minipage}
\end{figure*}

\subsection{Performance with baseline methods}
To validate the effectiveness of our method for improving one-step face restoration, we conduct a systematic comparison against our baseline methods. Specifically, we use Qwen-Image as the baseline model, fixing its conditional branch to a unified textual prompt and training it on the face dataset with the same number of iterations to ensure experimental fairness. As shown in Tab.~\ref{Table: compare baseline}, our method outperforms the baseline across most evaluation metrics, indicating that the proposed strategy enhances the overall performance of one-step face restoration across different architectures.

% \begin{table*}[t]
% \centering
% \caption{Quantitative comparison with the real-world datasets, the best and second-best results are highlighted in \best{Red} and \second{Blue}.}
% \label{table:real-data}
% \setlength{\tabcolsep}{5pt}
% \begin{tabular}{l|ccc|ccc}
% \toprule
% % \rowcolor{gray!15}
% \multirow{1}{*}{Methods} & \multicolumn{3}{c|}{CelebChild} & \multicolumn{3}{c}{LFW-Test} \\
% \cline{2-7}
% % \cmidrule(lr){2-5} \cmidrule(lr){6-9}
% % \rowcolor{gray!15}
% & Q-Align$\uparrow$ & LIQE$\uparrow$ & MUSIQ$\uparrow$
% & Q-Align$\uparrow$ & LIQE$\uparrow$ & MUSIQ$\uparrow$\\
% \midrule
% VQFR~\cite{gu2022vqfr} & 3.676 & \second{4.394} & 71.72 & 4.369 & \second{4.832} & 74.75\\

% RestoreFormer++~\cite{wang2023restoreformer++} & 3.594 &  4.077 & 69.40 & 4.077 & 4.471 & 72.25\\

% PGDiff~\cite{yang2023pgdiff} & 3.574 & 3.886 & 66.73 & 4.088 & 4.279 & 71.47 \\

% OSDFace~\cite{wang2025osdface} & \second{3.742} & 4.321 & \best{72.53} & \second{4.374} & \best{4.865} & \second{75.35}\\

% \midrule
% Ours (SD V2.1-base)
% & 3.584 & 3.963 & 68.71 & 4.236 & 4.727 & 73.42\\
% Ours (Qwen-Image)
% & \best{3.880} & \best{4.395} & \second{72.46} & \best{4.435} & 4.813 & \best{76.04} \\
% \bottomrule
% \end{tabular}
% \end{table*}

\subsection{Performance on CelebA-Test dataset}
\noindent\textbf{Quantitative Results.} As shown in Tab.~\ref{main:table}, our method surpasses competing approaches on most evaluation metrics. Specifically, across DISTS, LPIPS, and TOPIQ, our models based on two architectures consistently rank within the top two, indicating that guidance from high-dimensional representations effectively enhances detail preservation and perceptual consistency in reconstructed images. Moreover, the leading performance on Deg. and LMD further demonstrates our method’s superior ability to maintain facial identity consistency.

\begin{figure}[t]
\centering
\includegraphics[width=\linewidth]{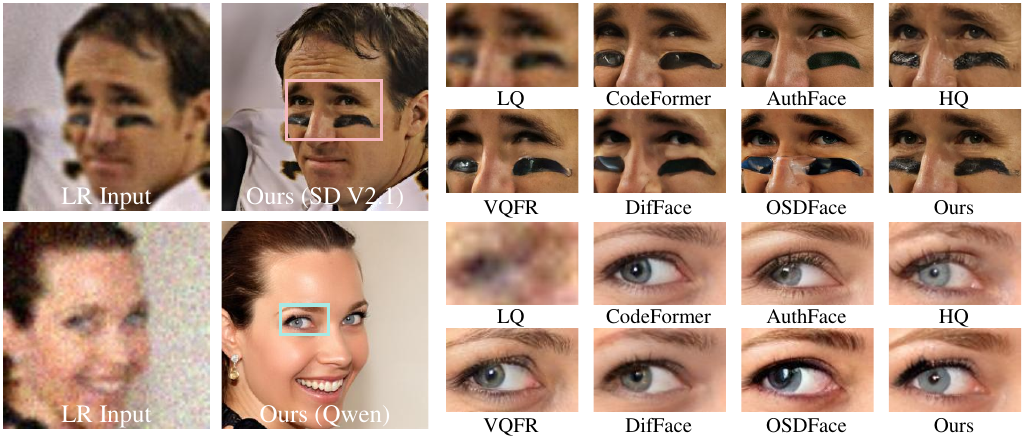}
\caption{Qualitative comparison on the CelebA-Test dataset.}
\label{f-visual_compare}
\vspace{-1em}
\end{figure}

\noindent\textbf{Qualitative Results.}
The visual comparisons in Fig.~\ref{f-visual_compare} show that the results generated by our method are the closest to the real images. Specifically, in the first-row example, compared with other methods, our method more accurately restores the details of the facial makeup region, with a glossiness that is closer to the GT image. In the second-row example, the other methods exhibit obvious deviations from the real image in both the color and shape of the pupils, whereas the restoration result produced by our method remains relatively consistent with the real image. These observations further demonstrate that the high-dimensional representations introduced in our method can effectively enhance the model’s ability to capture and restore fine facial details.

% The visual comparisons in Fig.~\ref{f-visual_compare} demonstrate that our method produces results that are the closest to the real images. Specifically, in the first-row example, affected by degradation, the other methods fail to properly attend to and restore the star-shaped makeup on the face, whereas only our method attempts to recover this makeup detail. In the fourth-row example, the other methods exhibit noticeable deviations from the ground truth in both the color and shape of the pupils, while our method yields a restoration result that remains relatively consistent with the ground truth. These observations further indicate that the high-dimensional representations we introduce can effectively enhance the model’s ability to capture and restore fine facial details.

\subsection{Performance on Real-World dataset}
As shown in Tab.~\ref{table:real-data}, our method trained based on Qwen-Image ranks first or second on most evaluation metrics, demonstrating its overall performance advantages. In contrast, our method trained on SD V2.1-base places greater emphasis on the fidelity of the reconstructed results, and therefore performs relatively weaker on no-reference evaluation metrics. Nevertheless, we further provide qualitative comparisons in the Appendix~\ref{appendix:a.2} to substantiate the practical effectiveness and advantages of the proposed method.

\subsection{Ablation Studies}
\begin{figure*}[t]
\centering
\begin{minipage}[t]{0.54\textwidth}
  \centering
  \includegraphics[width=\linewidth]{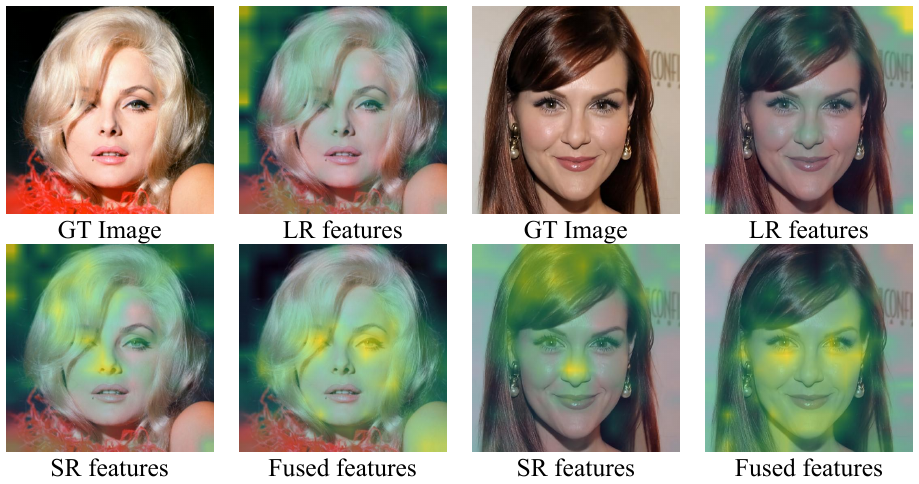}
  \caption{Heatmap visualizations of representation features.}
  \label{figure:hotmap}
\end{minipage}\hfill
\begin{minipage}[t]{0.42\textwidth}
  \centering
  \includegraphics[width=\linewidth]{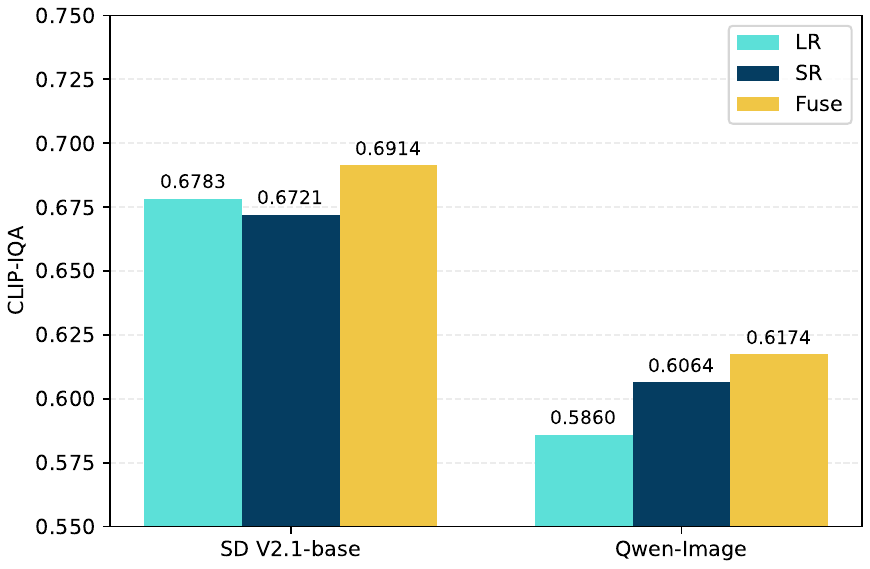}
  \caption{Ablation evaluation of different input images.}
  \label{abla:figure}
\end{minipage}
\vspace{-1em}
\end{figure*}

\noindent\textbf{Effectiveness of SDFM.} We validate the effectiveness of the proposed SDFM module through both qualitative visualization and quantitative experiments. As shown in Fig.~\ref{figure:hotmap}, we present heatmap visualizations of different features extracted by DINOv3. It can be observed that the LR features tend to focus on edge and structural regions of the image, whereas the SR features emphasize local details such as facial components and hair. In contrast, the features fused by SDFM attend more comprehensively to the entire face region, achieving a more balanced allocation of attention between global facial structure and local details. This indicates that the fused features are more suitable for the face restoration task.

In addition, we provide quantitative comparisons in Tab.~\ref{abla:table}. Specifically, we compare our full method with three alternative settings: using Text only, using DINO only, and using DINO together with SDFM. The results show that our method consistently outperforms these baselines across most evaluation metrics, thereby demonstrating the effectiveness of the SDFM module in improving face restoration performance.

\noindent\textbf{Ablation of LoRA rank.} We conduct comparative experiments under different LoRA rank settings to determine the optimal training configuration. As shown in Tab.~\ref{abla:table}, when the LoRA rank is set to 16, all restoration performance metrics achieve their best results. Therefore, in the actual training process, we fix the LoRA rank to 16.

\noindent\textbf{Ablation of DINO type.} To determine the most suitable pretrained representation model for face restoration, we conduct systematic ablation studies on different DINO types. Specifically, we evaluate three variants of DINOv3 and employ an additional MLP layer to align feature dimensions. As shown in Tab.~\ref{abla:table}, benefiting from its larger model capacity, DINOv3-L achieves the best performance across all evaluation metrics. Therefore, we adopt DINOv3-L as the representation extractor in subsequent experiments.

\noindent\textbf{Ablation of Input images.} We conduct ablation studies on the input configurations to verify the effectiveness of jointly incorporating SR and LR images for face restoration. As shown in Fig.~\ref{abla:figure}, the model achieves the best performance when both SR and LR image information are utilized simultaneously, indicating that the two sources provide complementary information. This result further validates the necessity and rationality of our method design.

\section{Discussion}
\noindent\textbf{Why are text prompts unsuitable for face restoration?} In image super-resolution, text prompts are commonly used as auxiliary guidance~\cite{wu2024seesr, fang2025one, wu2024one}. However, during the training of Qwen-Image, we observe that text prompts perform worse than visual features, as shown in Tab.~\ref{abla:table}. We attribute this to fundamental differences in information representation between the two tasks. In general image super-resolution, text prompts provide explicit semantic cues, such as object categories, attributes, or textual content, offering useful prior information for reconstruction. In contrast, face restoration relies heavily on identity-specific and fine-grained details that are difficult to describe precisely in language. Moreover, such discriminative textual information is hard to reliably infer from low-quality inputs. Consequently, text-based guidance provides limited and unstable constraints for facial details. High-dimensional visual representations, by contrast, encode facial structure and appearance more directly and at finer granularity, making them more suitable for guiding face restoration.

\noindent\textbf{The computational cost of our method.} 
Although our framework introduces an additional forward pass of an off-the-shelf restorer and high-dimensional feature extraction, we argue that this trade-off (quality over speed) is highly worthwhile for identity-critical face restoration under severe degradations. Furthermore, since both the intermediate restorer and the final generative model operate in a one-step manner, the overall pipeline successfully bypasses the heavy iterative sampling (e.g., 50 steps) of standard multi-step diffusion models, maintaining a highly competitive inference latency. We further report the computational consumption and runtime of our method in the Appendix~\ref{appendix:a.3}.

\definecolor{retouchbg}{RGB}{245,226,204}
\definecolor{editbg}{RGB}{206,210,255}  
\begin{table*}[t]
\centering
\small 
\caption{Ablation studies of our method. The left table compares the proposed method with three ablated variants: using Text only, using DINO only, and using DINO together with SDFM. The right table evaluates the impact of varying the LoRA rank and the DINO module type on performance.}
\label{abla:table}
\setlength{\tabcolsep}{4pt}
\begin{subtable}[t]{0.60\textwidth}
\centering
\begin{tabular}{ccc|cccc}
\Xhline{1.1pt}
 DINO & SDFM & Text & C-IQA$\uparrow$  & Q-Align$\uparrow$ & LIQE$\uparrow$ & Deg.$\downarrow$\\
\hline
\rowcolor{retouchbg} 
\multicolumn{7}{l}{\textit{SD V2.1-base}}\\
\xmark & \xmark & \checkmark & \textbf{0.6947} & 4.522 & 4.869 & 30.52 \\
\checkmark & \xmark & \xmark & 0.6721 & 4.341 & 4.741 & 30.39\\
\checkmark & \checkmark & \xmark & 0.6914 & \textbf{4.544} & \textbf{4.882} & \textbf{30.12}\\
\hline
\rowcolor{editbg} 
\multicolumn{7}{l}{\textit{Qwen-Image}}\\
\xmark & \xmark & \checkmark & 0.5860 & 4.391 & 4.710 & 34.56 \\
\checkmark & \xmark & \xmark & 0.6064 & 4.495 & 4.807 & 33.42 \\
\checkmark & \checkmark & \xmark & \textbf{0.6174} & \textbf{4.546} & \textbf{4.841} & \textbf{33.30}\\
\Xhline{1.1pt}
\end{tabular}
\end{subtable}
\hfill
\begin{subtable}[t]{0.37\textwidth}
\centering
\begin{tabular}{c|ccc}
\Xhline{1.1pt}
 & C-IQA$\uparrow$ & Q-Align$\uparrow$ & LIQE$\uparrow$ \\
\hline
\rowcolor{retouchbg} 
\multicolumn{4}{l}{\textit{LoRA Rank}}\\
 16 & \textbf{0.6914} & \textbf{4.544} & \textbf{4.882}\\
 32 & 0.6864 & 4.429 & 4.785\\
 64 & 0.6782 & 4.470 & 4.872\\
\hline
\rowcolor{editbg} 
\multicolumn{4}{l}{\textit{DINO type}}\\
L & \textbf{0.6914} & \textbf{4.544} & \textbf{4.882} \\
B & 0.6843 & 4.431 & 4.814 \\
S & 0.6779 & 4.426 & 4.771 \\
\Xhline{1.1pt}
\end{tabular}
\end{subtable}
\end{table*}

\section{Limitations}
Although HDRFace consistently improves performance across different generative backbones, it still has several limitations. First, the additional intermediate restoration model introduces extra computational overhead and increases inference cost. Second, since our method relies on the intermediate result for structural guidance, the final restoration quality can be affected by its reliability. Although SDFM adaptively fuses and selects high-dimensional representations to reduce the impact of misleading features, using only a single intermediate result may still be insufficient under extremely severe degradations or complex real-world scenarios. A more robust alternative could be to filter or adaptively select from multiple candidate SR results at test time. We leave this test-time selection strategy for future work to further improve robustness and generalization.

\section{Conclusion}
In this paper, we presented HDRFace, a novel framework that addresses the ill-posed nature of the face restoration task by injecting high-dimensional representation into the conditional generation process. Recognizing the limitations of relying solely on low-quality inputs, we leveraged the high-dimensional representation encoder to extract rich, fine-grained representations from both the input and intermediate restoration results. To effectively integrate these priors, we introduced SDFM, an adaptive fusion mechanism that decouples structure from details, ensuring global consistency while enhancing local fidelity. Extensive experiments on SD V2.1-base and Qwen-Image demonstrate that our method consistently improves restoration quality and identity preservation across different architectures. Our work not only offers a robust solution for severe degradation scenarios but also highlights the potential of high-dimensional representation guidance in conditional generative tasks, pointing to a promising direction for future research in face restoration task.
\bibliographystyle{unsrt}
\bibliography{neurips_2026.bib}

%%%%%%%%%%%%%%%%%%%%%%%%%%%%%%%%%%%%%%%%%%%%%%%%%%%%%%%%%%%%
\appendix
%%%%%%%%%%%%%%%%%%%%%%%%%%%%%%%%%%%%%%%%%%%%%%%%%%%%%%%%%%%%
\clearpage
\section{Appendix}

\subsection{Backbone Architectures and High-Dimensional Representation Injection}
\label{appendix:backbone_injection}

This section clarifies how the proposed high-dimensional representation condition is injected into the two generative backbones used in HDRFace. After extracting the DINOv3 representations~\cite{simeoni2025dinov3} and applying SDFM, we map the fused representation to the conditioning space required by each backbone:
\begin{equation}
\mathbf{C}_{\mathrm{HDR}}=\phi(\mathbf{F}_{\mathrm{fused}}),
\end{equation}
where $\phi$ denotes a lightweight projection module that aligns $\mathbf{F}_{\mathrm{fused}}$ with the backbone-specific conditional feature dimension. This design keeps the generative backbone unchanged and only replaces the semantic condition passed through its original text-conditioning interface.

\textbf{Injection into SD V2.1-base.}
SD V2.1-base follows the latent diffusion paradigm~\cite{rombach2022high}, where images are processed in the VAE latent space and a denoising U-Net receives OpenCLIP text features through cross-attention. In the original text-to-image model, the text encoder produces \texttt{text\_embedding}, which is then used as the cross-attention condition of the U-Net. In HDRFace, we directly replace this \texttt{text\_embedding} input with $\mathbf{C}_{\mathrm{HDR}}$. Therefore, the high-dimensional facial representation enters SD V2.1-base through the same cross-attention pathway as textual semantics, enabling representation-conditioned restoration without introducing an additional conditioning branch.

\textbf{Injection into Qwen-Image.}
Qwen-Image is a 20B MMDiT image foundation model. In our restoration setting, we follow the Rectified Flow formulation~\cite{liu2022flow} and the conditional training strategy used in ODTSR~\cite{fang2025one}. Different from SD V2.1-base, Qwen-Image uses a transformer-based denoising backbone and accepts semantic guidance through a text-conditioning stream. In HDRFace, $\mathbf{C}_{\mathrm{HDR}}$ is fed as the \texttt{text\_condition} of Qwen-Image. The noisy LR latent condition $\mathbf{z}_{\mathrm{cond}}$ remains an auxiliary image-space condition and is kept separate from the high-dimensional semantic condition; its construction is described in Appendix~\ref{appendix:a.1}.

\subsection{Training Details of the Overall Restoration Model}
\label{appendix:a.1}

In this section, we provide a detailed description of the training process of our one-step face restoration model under two different generative architectures, namely SD V2.1-base and Qwen-Image. We elaborate on the construction of the noisy latent input, the design of the conditional inputs, and the formulation of each loss term in the unified training objective.

\subsubsection{Training Process under SD V2.1-base}
For SD V2.1-base, we treat the low-resolution image as the noisy input to the diffusion model and predefine a fixed timestep $T$. Conditioned on the fused high-dimensional facial representation $\mathbf{F}_{\mathrm{fused}}$, the model performs one-step denoising to generate a high-quality restored face image. The process can be formulated as follows:
\begin{equation}
\hat{\mathbf{I}}_{\mathrm{face}}=\mathcal{D}_{\mathrm{VAE}}\!\left(\frac{\mathbf{z}_{T}-\sqrt{1-\bar{\alpha}_{T}}\,\epsilon_{\theta}\!\left(\mathbf{z}_{T},T,\mathbf{F}_{\mathrm{fused}}\right)}{\sqrt{\bar{\alpha}_{T}}}\right),
\end{equation}
where $T$ is a fixed timestep in our implementation, $\mathbf{z}_{T}$ denotes the LR image in the latent space, and $\bar{\alpha}_{T}$ denotes the cumulative noise schedule coefficient at timestep $T$.

After obtaining the predicted image $\hat{\mathbf{I}}_{\mathrm{face}}$, we compute the training loss following~\cite{wang2025osdface}, which can be formulated as:
\begin{equation}
\begin{aligned}
\mathcal{L}_{\mathrm{total}} \;=\;&
\lambda_{\mathrm{G}}\,\mathcal{L}_{G}\!\left(\mathbf{z}_{\mathrm{HR}}, \hat{\mathbf{z}}_{\mathrm{face}}\right)
+\lambda_{\mathrm{ID}}\,\mathcal{L}_{\mathrm{ID}}\!\left(\mathbf{I}_{\mathrm{HR}}, \hat{\mathbf{I}}_{\mathrm{face}}\right) \\
&+\lambda_{\mathrm{per}}\,\mathcal{L}_{\mathrm{per}}\!\left(\mathbf{I}_{\mathrm{HR}}, \hat{\mathbf{I}}_{\mathrm{face}}\right)
+\lambda_{\mathrm{rec}}\,\mathrm{MSE}\!\left(\mathbf{I}_{\mathrm{HR}}, \hat{\mathbf{I}}_{\mathrm{face}}\right),
\end{aligned}
\end{equation}
where $\hat{\mathbf{z}}_{\mathrm{face}}$ and $\mathbf{z}_{\mathrm{HR}}$ represent the predicted image and the ground-truth image in the latent space, respectively. Each loss term is described in detail below.

\textbf{Adversarial Loss.} To facilitate training and enhance the realism of the generated results, we employ a GAN loss $\mathcal{L}_{G}$, which is defined as:
\begin{equation}
\begin{gathered}
\mathcal{L}_{D} =
-\mathbb{E}\!\left[\log D\!\left(\mathbf{z}_{\mathrm{HR}}\mid \mathbf{F}_{\mathrm{fused}}\right)\right]
-\mathbb{E}\!\left[\log\!\left(1-D\!\left(\hat{\mathbf{z}}_{\mathrm{face}}\mid \mathbf{F}_{\mathrm{fused}}\right)\right)\right], \\
\mathcal{L}_{G} =
-\mathbb{E}\!\left[\log D\!\left(\hat{\mathbf{z}}_{\mathrm{face}}\mid \mathbf{F}_{\mathrm{fused}}\right)\right].
\end{gathered}
\label{eq:adversarial_loss_sd}
\end{equation}
Here, the discriminator $D$ is conditioned on the fused facial representation $\mathbf{F}_{\mathrm{fused}}$ to provide stronger guidance during adversarial training.

\textbf{Identity Loss.} To preserve facial identity during restoration, we employ a pretrained ArcFace model to extract identity embeddings from both the restored image and the HR ground truth, and compute their cosine similarity. Accordingly, the identity loss $\mathcal{L}_{\mathrm{ID}}$ is formulated as:
\begin{equation}
\mathcal{L}_{\mathrm{ID}} = 1 - \cos\!\big( Arc(\mathbf{I}_{\mathrm{HR}}),\, Arc(\hat{\mathbf{I}}_{\mathrm{face}}) \big),
\end{equation}
where $Arc(\cdot)$ denotes the identity embedding extracted by the pretrained ArcFace network.

\textbf{Perceptual Loss.} For the perceptual loss $\mathcal{L}_{\mathrm{per}}$, we replace the commonly used LPIPS with DISTS to better capture texture details. In addition, we incorporate edge-aware perceptual evaluation by applying the Sobel operator to extract edge information. Our formulation is given by:
\begin{equation}
\mathcal{L}_{\mathrm{per}}(\mathbf{I}_{\mathrm{HR}}, \hat{\mathbf{I}}_{\mathrm{face}}) = \mathcal{L}_{\mathrm{DISTS}}(\mathbf{I}_{\mathrm{HR}}, \hat{\mathbf{I}}_{\mathrm{face}}) + \mathcal{L}_{\mathrm{DISTS}}\!\big( S(\mathbf{I}_{\mathrm{HR}}),\, S(\hat{\mathbf{I}}_{\mathrm{face}}) \big),
\end{equation}
where $S(\cdot)$ denotes the Sobel operator, which extracts edge information to compute the edge-aware DISTS.

\textbf{Reconstruction Loss.} The reconstruction loss $\mathcal{L}_{\mathrm{rec}}$ is implemented as the standard mean squared error (MSE) between the restored image and the HR ground truth in the pixel space:
\begin{equation}
\mathcal{L}_{\mathrm{rec}} = \mathrm{MSE}(\mathbf{I}_{\mathrm{HR}}, \hat{\mathbf{I}}_{\mathrm{face}}).
\end{equation}

\subsubsection{Training Process under Qwen-Image}
Unlike the training strategy used in SD V2.1-base, when training Qwen-Image, we first introduce a certain degree of noise perturbation to the low-quality images and, following the formulation of Rectified Flow, adopt a linear noising and denoising process, which can be expressed as:
\begin{equation} 
\begin{gathered}
\mathbf{z}_T = (1 - \sigma_T)\,\mathbf{z}_{\mathrm{LR}} + \sigma_T\,\boldsymbol{\epsilon},\quad \boldsymbol{\epsilon} \sim \mathcal{N}(\mathbf{0}, \mathbf{I}), \\
\hat{\mathbf{I}}_{\mathrm{face}} =
\mathcal{D}_{\mathrm{VAE}}\!\left(
\mathbf{z}_T - \sigma_T\,\mathbf{v}_{\theta}\!\left(\mathbf{z}_T, T, \mathbf{z}_{\mathrm{cond}}, \mathbf{F}_{\mathrm{fused}}\right)
\right),
\end{gathered}
\label{eq:qwen_image_process}
\end{equation}
where $\mathbf{v}_{\theta}$ denotes the velocity prediction network in the Rectified Flow framework, and $\sigma_T$ is the noise intensity at timestep $T$.

\textbf{Conditional Input Design.} During training, following the setting of ODTSR~\cite{fang2025one}, we further introduce the low-quality image with injected random-intensity noise, denoted as $\mathbf{z}_{\mathrm{cond}}$, into the denoising network as an additional conditional input. Specifically, $\mathbf{z}_{\mathrm{cond}}$ is formulated as:
\begin{equation}
\mathbf{z}_{\mathrm{cond}} = (1 - \sigma_{t'})\,\mathbf{z}_{\mathrm{LR}} + \sigma_{t'}\,\boldsymbol{\epsilon}, \quad \sigma_{t'} = (1 - f) \cdot \sigma_{T}, \quad \boldsymbol{\epsilon} \sim \mathcal{N}(\mathbf{0}, \mathbf{I}),  
\end{equation}
where the variable $f$ is randomly sampled from the interval $[0,1]$. By using input images with varying noise intensities as a modulation signal, the model is able to achieve stronger editing capability under high noise levels, while preserving higher content fidelity when the noise intensity is low.

Furthermore, with probability $p_{\mathrm{clean}}$, the above noise augmentation process is skipped, and $\mathbf{z}_{\mathrm{cond}}$ is directly set to $\mathbf{z}_{\mathrm{LR}}$, i.e., the original low-quality latent is used as the conditional input. This design enables the model to maintain robustness to degraded images while preserving high-fidelity reconstruction capability under noise-free conditions, which helps stabilize perceptual quality metrics during training. This can be formally expressed as:
\begin{equation}
\tilde{\mathbf{z}}_{\mathrm{LR}} = 
\begin{cases} 
(1 - \sigma_{t'})\,\mathbf{z}_{\mathrm{LR}} + \sigma_{t'}\,\boldsymbol{\epsilon}, & u \geq p_{\mathrm{clean}}, \\ 
\mathbf{z}_{\mathrm{LR}}, & u < p_{\mathrm{clean}},
\end{cases}
\end{equation}
where $u \sim \mathcal{U}(0,1)$ is a uniformly sampled random variable.

\textbf{Training Objective.} For the training of Qwen-Image, the overall optimization objective remains consistent with the unified formulation in the main text, but the specific implementations of the reconstruction, perceptual, and adversarial loss terms differ from those used in SD V2.1-base. Specifically, we adopt MSE loss as the reconstruction loss, LPIPS loss as the perceptual loss, and further incorporate a GAN loss to enhance the realism of the generated results:
\begin{equation}
\begin{aligned}
\mathcal{L}_{\mathrm{total}} \;=\;& \lambda_{\mathrm{rec}}\,\mathrm{MSE}\!\left(\mathbf{I}_{\mathrm{HR}}, \hat{\mathbf{I}}_{\mathrm{face}}\right) \\
&+ \lambda_{\mathrm{per}}\, \mathcal{L}_{\mathrm{LPIPS}}\!\left(\mathbf{I}_{\mathrm{HR}}, \hat{\mathbf{I}}_{\mathrm{face}}\right) \\
&+ \lambda_{\mathrm{G}}\, \mathcal{L}_{G}\!\left(\mathbf{z}_{\mathrm{HR}}, \hat{\mathbf{z}}_{\mathrm{face}}\right).
\end{aligned}
\end{equation}

\textbf{Adversarial Loss for Qwen-Image.} Different from the standard GAN loss used in SD V2.1-base, for Qwen-Image, we adopt a relativistic average GAN loss formulation, which has been shown to provide more stable adversarial training under the Rectified Flow framework. The corresponding GAN loss is formulated as:
\begin{equation}
\mathcal{L}_{G} = \frac{1}{2}\Big[\mathbb{E}\big[\log(1 - \sigma(D(\mathbf{z}_{\mathrm{HR}}) - \mathbb{E}[D(\hat{\mathbf{z}}_{\mathrm{face}})]))\big] + \mathbb{E}\big[\log \sigma(D(\hat{\mathbf{z}}_{\mathrm{face}}) - \mathbb{E}[D(\mathbf{z}_{\mathrm{HR}})])\big]\Big],
\end{equation}
where $\sigma(\cdot)$ denotes the sigmoid function, and $D(\cdot)$ is the discriminator network.

\begin{table*}[t]
\centering
\small
\caption{Full training and inference details.}
\label{tab:full_training_details}
\setlength{\tabcolsep}{4pt}
\renewcommand{\arraystretch}{1.1}
\begin{tabular}{l|cc}
\toprule
Item & SD V2.1-base & Qwen-Image \\
\midrule
Training dataset & FFHQ & FFHQ \\
Image resolution & $512 \times 512$ & $512 \times 512$ \\
Degradation pipeline & VQFR pipeline & VQFR pipeline \\
Representation encoder & DINOv3-L & DINOv3-L \\
Projection hidden dimension & $1024 \to 1024$ &$1024 \to 3584$  \\
Optimizer & AdamW  & AdamW \\
Learning rate & 2e-4 & 5e-5 \\
Batch size & 2 & 1 \\
Training iterations & 100k & 10k \\
Trainable parameters & LoRA & Full fine-tuning \\
LoRA rank / alpha & 16 / 16 & N/A \\
Fixed timestep $T$ & 399 & 750 \\
$p_{\mathrm{clean}}$ & N/A & 0.5 \\
Loss weights & $\lambda_{\mathrm{G}}, \lambda_{\mathrm{ID}}, \lambda_{\mathrm{per}}, \lambda_{\mathrm{rec}}$: 5e-3, 0.5, 1.0, 2.0 & $\lambda_{\mathrm{G}}, \lambda_{\mathrm{per}}, \lambda_{\mathrm{rec}}$: 0.02, 1.0, 1.0 \\
Discriminator architecture & SDXL patch GAN & Wan-2.1 \\
Intermediate restorer & Qwen-Image & SD V2.1-base \\
\bottomrule
\end{tabular}
\end{table*}

\subsubsection{Summary}
Through the above training strategies, we complete the training of one-step face restoration models based on different generative architectures (i.e., SD V2.1-base and Qwen-Image), while incorporating high-dimensional facial representations into the model learning process. Despite the architectural differences and the corresponding adjustments in the noising/denoising formulation, conditional input design, and specific loss implementations, the overall training objective remains consistent with the unified framework presented in the main text, thereby achieving high-fidelity and high-quality face restoration across both architectures.

\subsection{Full Training and Inference Details}
\label{appendix:training_details_table}
Tab.~\ref{tab:full_training_details} summarizes the implementation details needed to reproduce the two HDRFace variants.

\subsection{More Qualitative comparison}
\label{appendix:a.2}
\begin{figure*}[t]
\centering
\setlength{\tabcolsep}{2pt} % 水平间距
\renewcommand{\arraystretch}{0.0} % 垂直紧凑
\footnotesize % 标注字号

% 行1
\begin{tabular}{cccccccc}
\begin{minipage}{0.130\linewidth}\centering
\includegraphics[width=\linewidth]{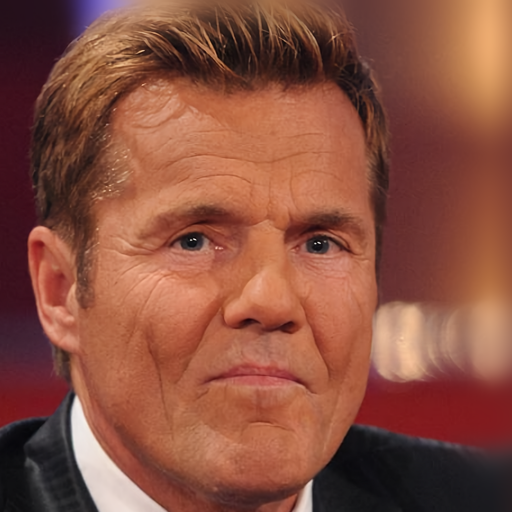}\
HQ
\end{minipage} &
% \begin{minipage}{0.124\linewidth}\centering
% \includegraphics[width=\linewidth]{images/00000040_LR.png}\
% LQ
% \end{minipage} &
\begin{minipage}{0.130\linewidth}\centering
\includegraphics[width=\linewidth]{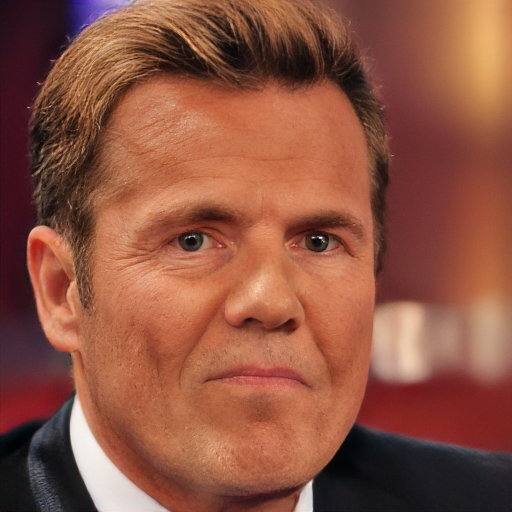}\
VQFR
\end{minipage} &
\begin{minipage}{0.130\linewidth}\centering
\includegraphics[width=\linewidth]{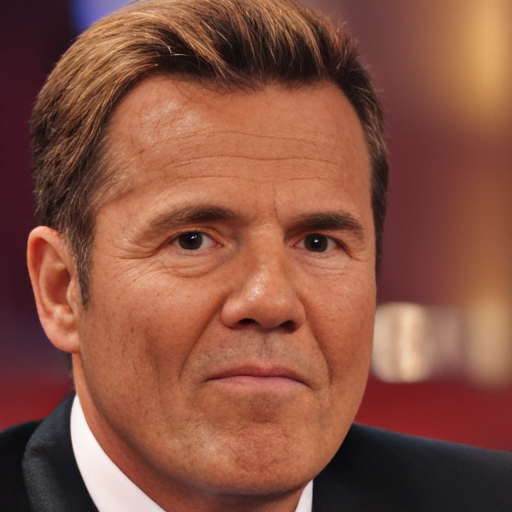}\
CodeF
\end{minipage} &
\begin{minipage}{0.130\linewidth}\centering
\includegraphics[width=\linewidth]{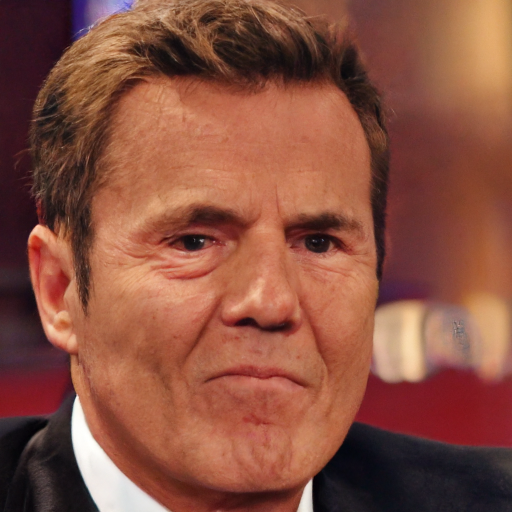}\
DifFace
\end{minipage} &
\begin{minipage}{0.130\linewidth}\centering
\includegraphics[width=\linewidth]{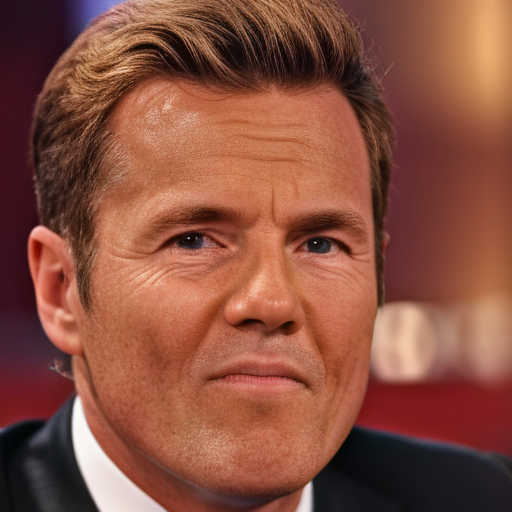}\
Authface
\end{minipage} &
\begin{minipage}{0.130\linewidth}\centering
\includegraphics[width=\linewidth]{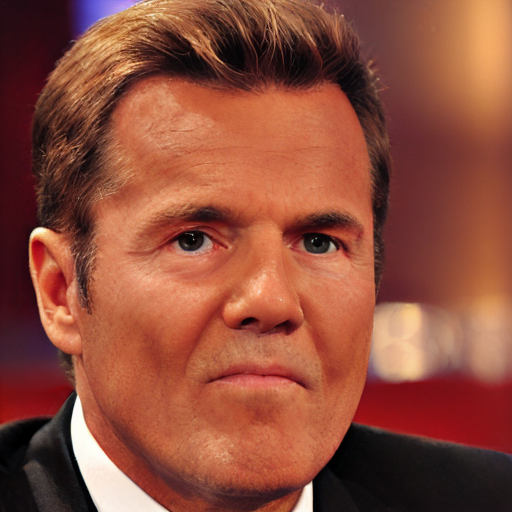}\
OSDFace
\end{minipage} &
\begin{minipage}{0.130\linewidth}\centering
\includegraphics[width=\linewidth]{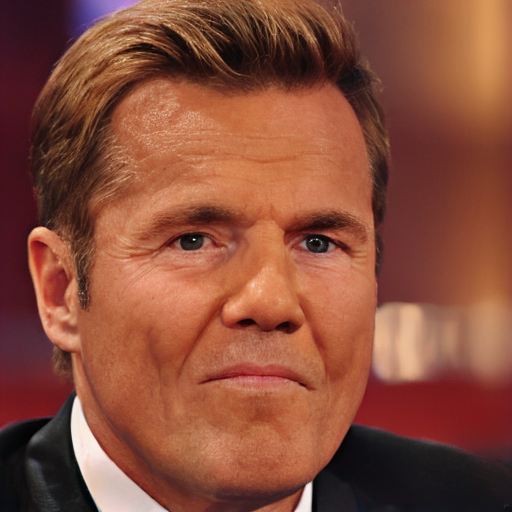}\
Ours
\end{minipage}
\end{tabular}

\vspace{2pt}

% 行2
\begin{tabular}{ccccccc}
\begin{minipage}{0.130\linewidth}\centering
\includegraphics[width=\linewidth]{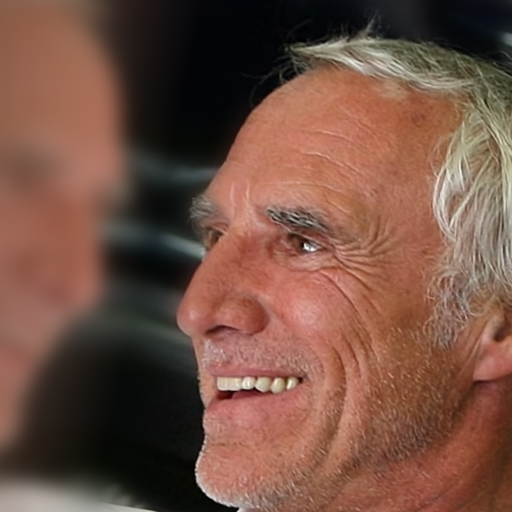}\
HQ
\end{minipage} &
% \begin{minipage}{0.124\linewidth}\centering
% \includegraphics[width=\linewidth]{images/00000044_LR.png}\
% LQ
% \end{minipage} &
\begin{minipage}{0.130\linewidth}\centering
\includegraphics[width=\linewidth]{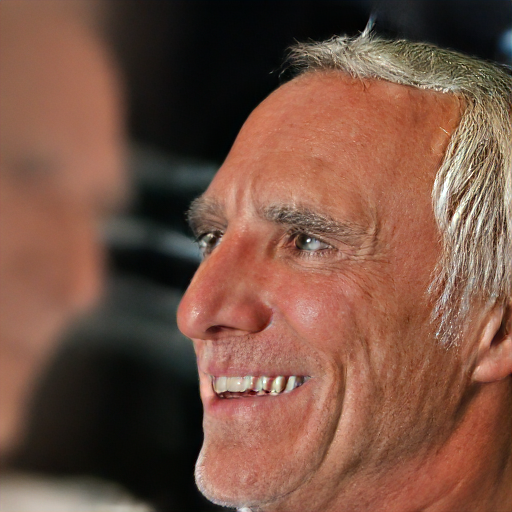}\
VQFR
\end{minipage} &
\begin{minipage}{0.130\linewidth}\centering
\includegraphics[width=\linewidth]{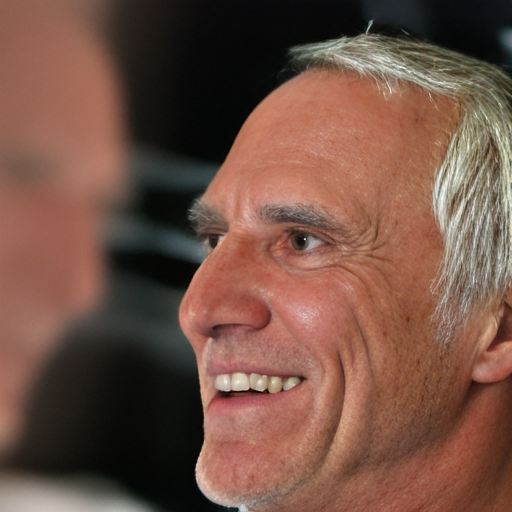}\
CodeF
\end{minipage} &
\begin{minipage}{0.130\linewidth}\centering
\includegraphics[width=\linewidth]{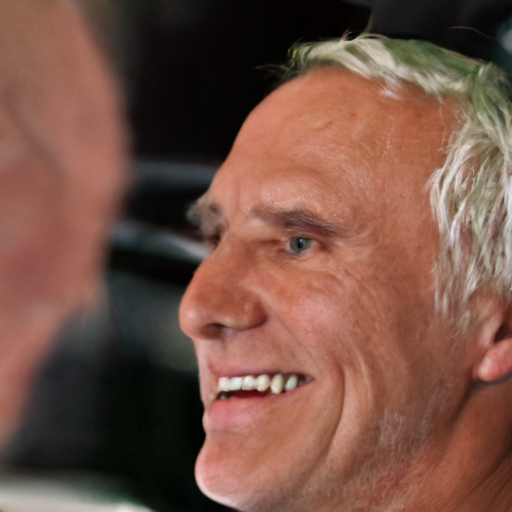}\
DifFace
\end{minipage} &
\begin{minipage}{0.130\linewidth}\centering
\includegraphics[width=\linewidth]{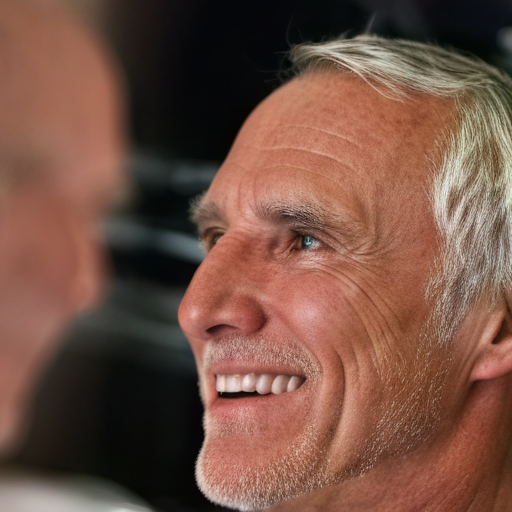}\
Authface
\end{minipage} &
\begin{minipage}{0.130\linewidth}\centering
\includegraphics[width=\linewidth]{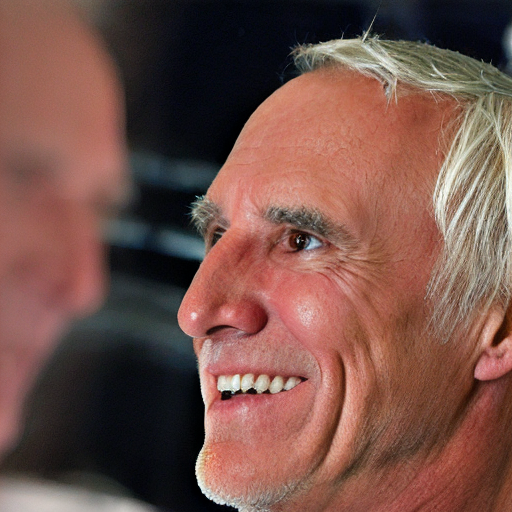}\
OSDFace
\end{minipage} &
\begin{minipage}{0.130\linewidth}\centering
\includegraphics[width=\linewidth]{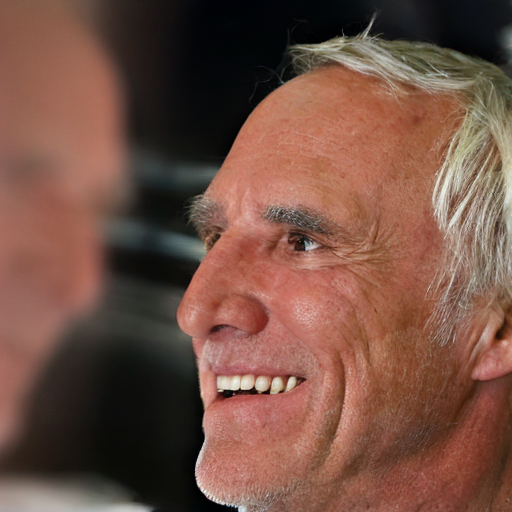}\
Ours
\end{minipage}
\end{tabular}

\vspace{2pt}

% 行3
\begin{tabular}{ccccccc}
\begin{minipage}{0.130\linewidth}\centering
\includegraphics[width=\linewidth]{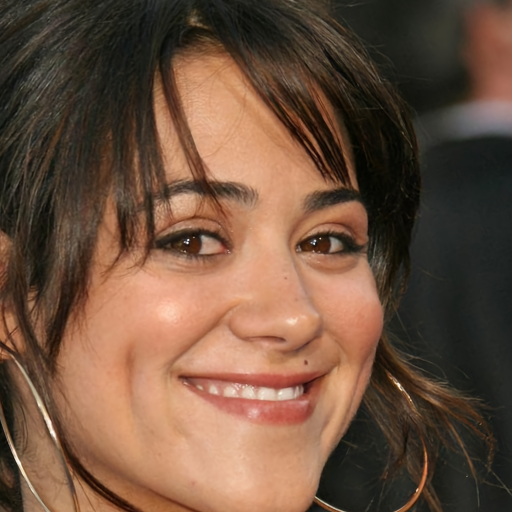}\
HQ
\end{minipage} &
% \begin{minipage}{0.124\linewidth}\centering
% \includegraphics[width=\linewidth]{images/00000057_LR.png}\
% LQ
% \end{minipage} &
\begin{minipage}{0.130\linewidth}\centering
\includegraphics[width=\linewidth]{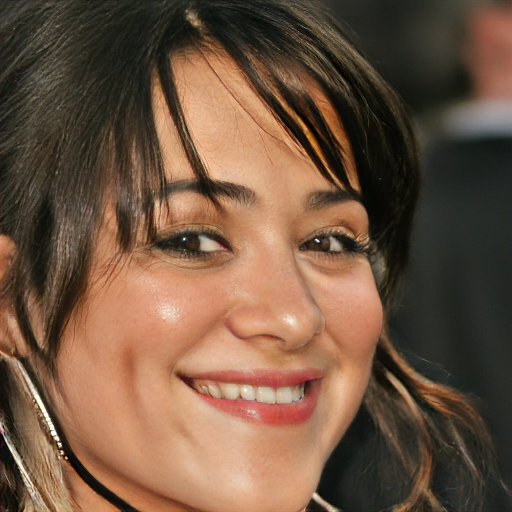}\
VQFR
\end{minipage} &
\begin{minipage}{0.130\linewidth}\centering
\includegraphics[width=\linewidth]{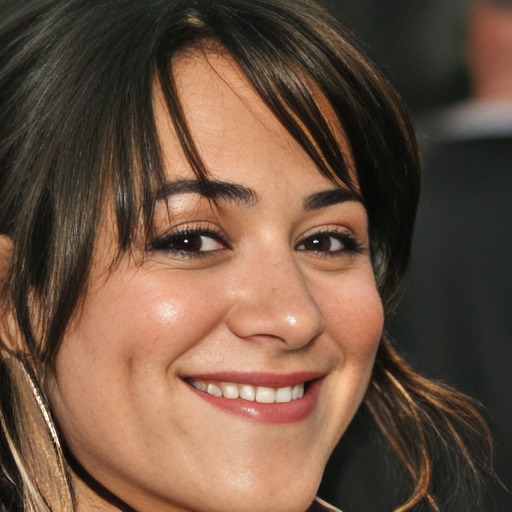}\
CodeF
\end{minipage} &
\begin{minipage}{0.130\linewidth}\centering
\includegraphics[width=\linewidth]{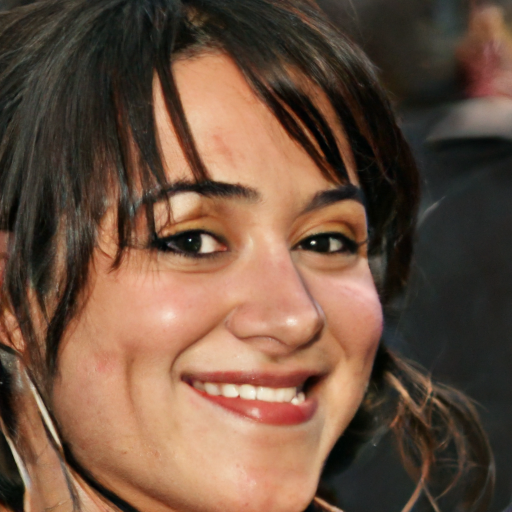}\
DifFace
\end{minipage} &
\begin{minipage}{0.130\linewidth}\centering
\includegraphics[width=\linewidth]{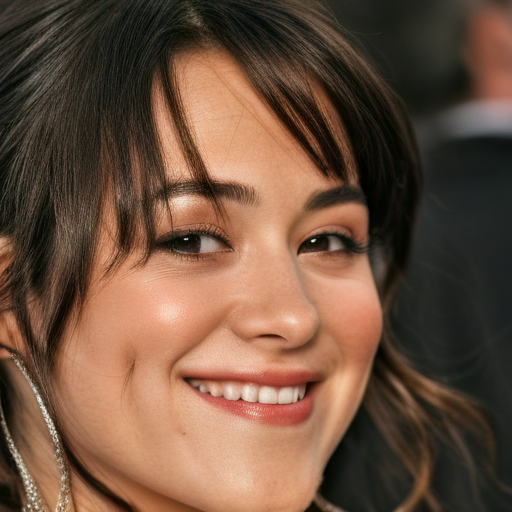}\
Authface
\end{minipage} &
\begin{minipage}{0.130\linewidth}\centering
\includegraphics[width=\linewidth]{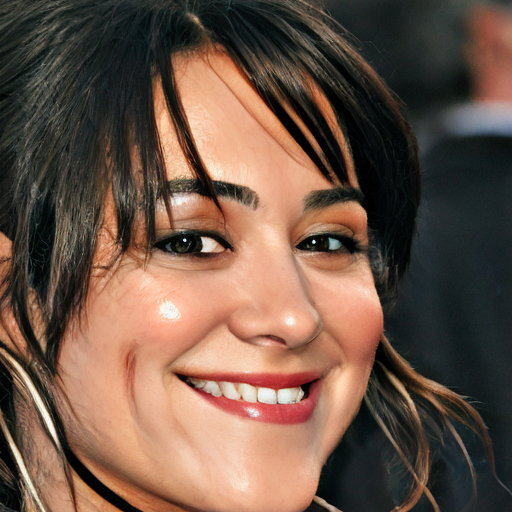}\
OSDFace
\end{minipage} &
\begin{minipage}{0.130\linewidth}\centering
\includegraphics[width=\linewidth]{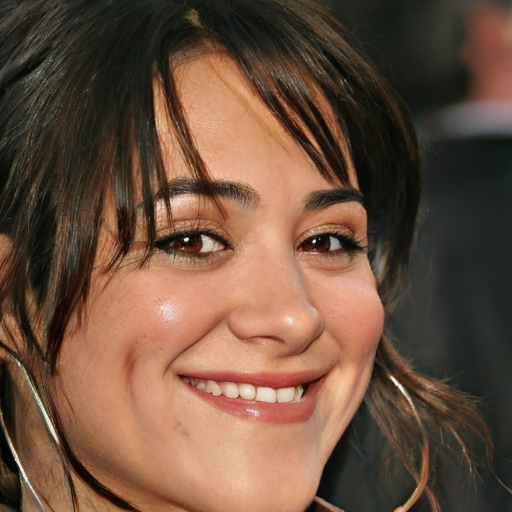}\
Ours
\end{minipage}
\end{tabular}

\vspace{2pt}

\begin{tabular}{ccccccc}
\begin{minipage}{0.130\linewidth}\centering
\includegraphics[width=\linewidth]{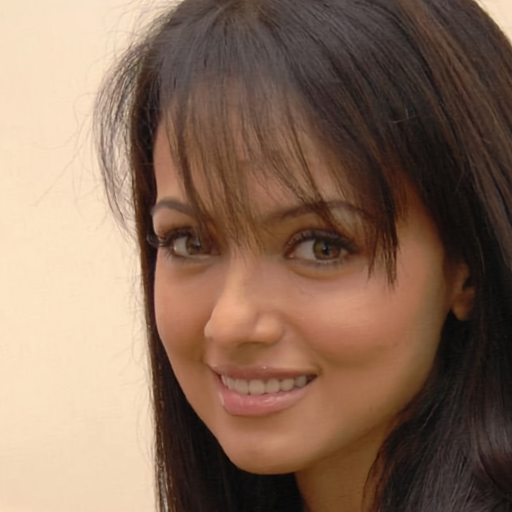}\
HQ
\end{minipage} &
% \begin{minipage}{0.125\linewidth}\centering
% \includegraphics[width=\linewidth]{images/00000076_LR.png}\
% LQ
% \end{minipage} &
\begin{minipage}{0.130\linewidth}\centering
\includegraphics[width=\linewidth]{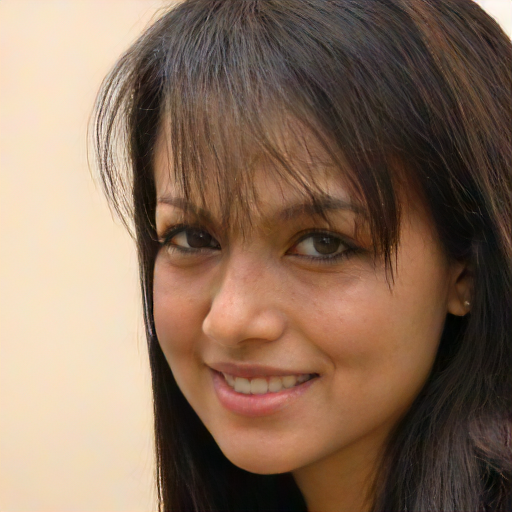}\
VQFR
\end{minipage} &
\begin{minipage}{0.130\linewidth}\centering
\includegraphics[width=\linewidth]{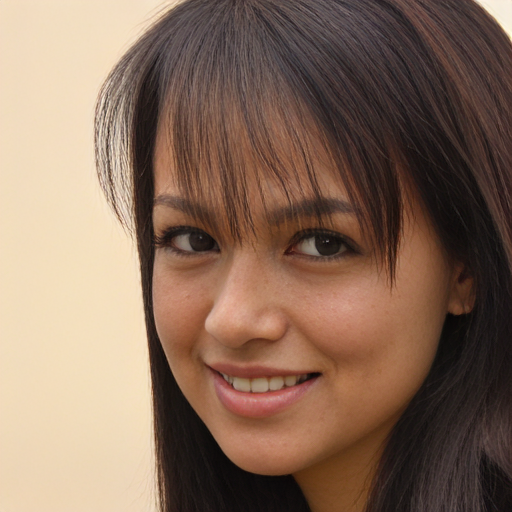}\
CodeF
\end{minipage} &
\begin{minipage}{0.130\linewidth}\centering
\includegraphics[width=\linewidth]{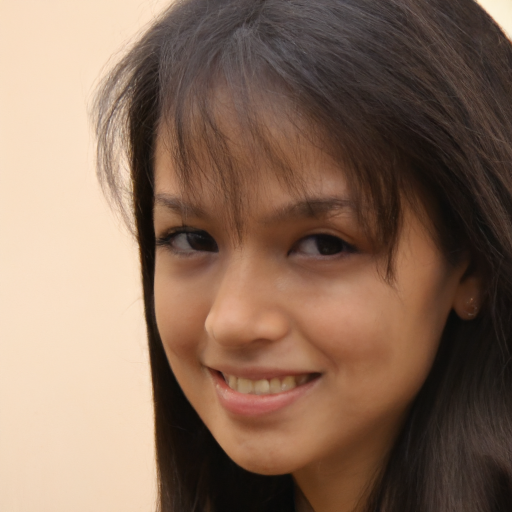}\
DifFace
\end{minipage} &
\begin{minipage}{0.130\linewidth}\centering
\includegraphics[width=\linewidth]{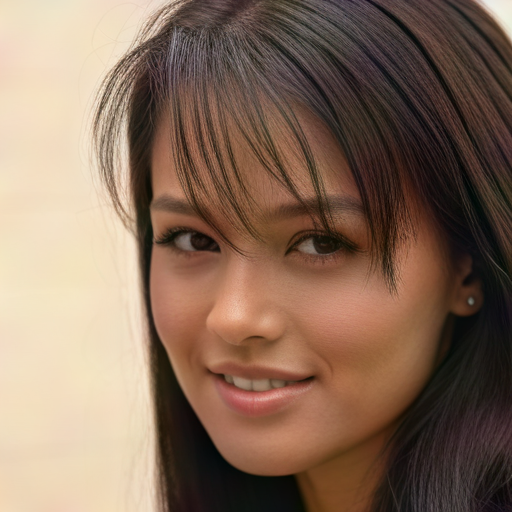}\
Authface
\end{minipage} &
\begin{minipage}{0.130\linewidth}\centering
\includegraphics[width=\linewidth]{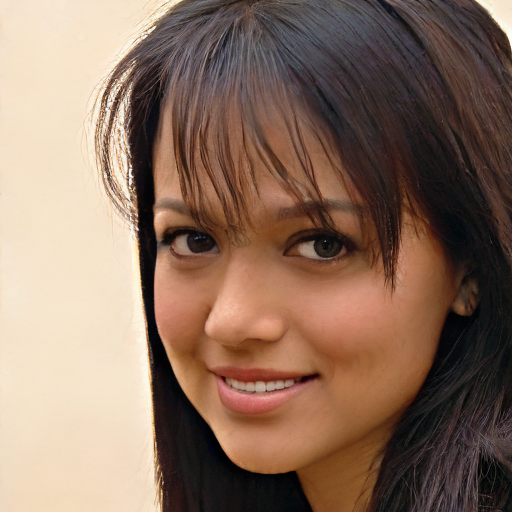}\
OSDFace
\end{minipage} &
\begin{minipage}{0.130\linewidth}\centering
\includegraphics[width=\linewidth]{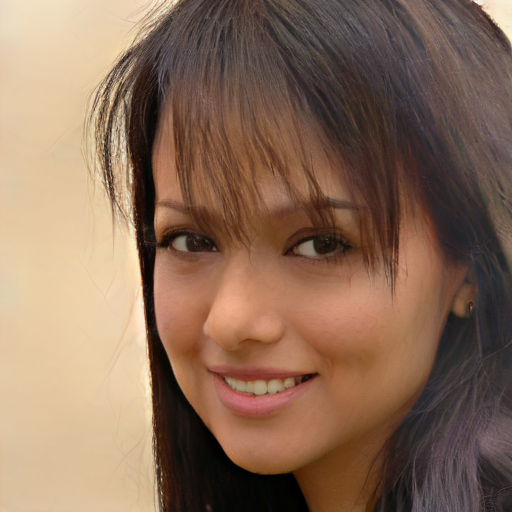}\
Ours
\end{minipage}
\end{tabular}

\vspace{2pt}

\begin{tabular}{ccccccc}
\begin{minipage}{0.130\linewidth}\centering
\includegraphics[width=\linewidth]{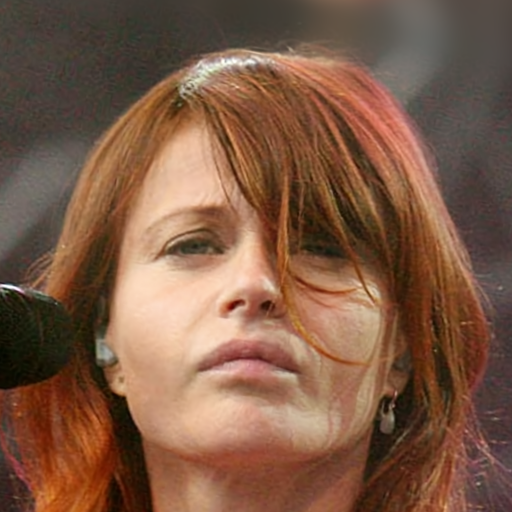}\
HQ
\end{minipage} &
% \begin{minipage}{0.125\linewidth}\centering
% \includegraphics[width=\linewidth]{images/00000141_LR.png}\
% LQ
% \end{minipage} &
\begin{minipage}{0.130\linewidth}\centering
\includegraphics[width=\linewidth]{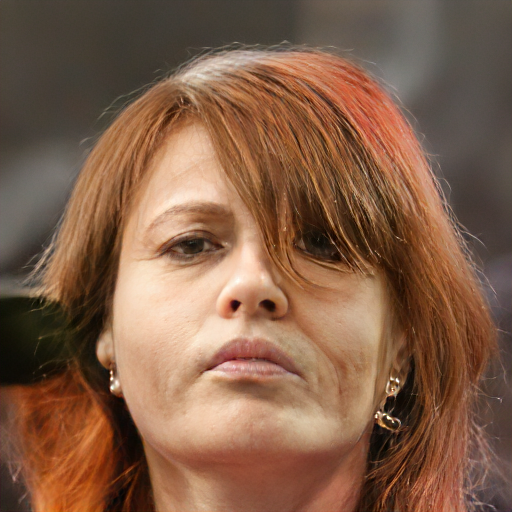}\
VQFR
\end{minipage} &
\begin{minipage}{0.130\linewidth}\centering
\includegraphics[width=\linewidth]{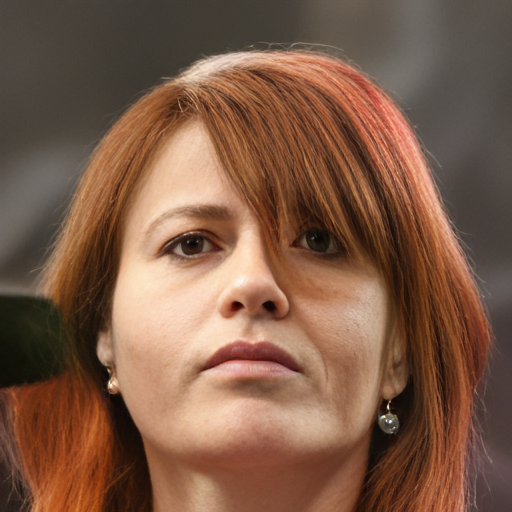}\
CodeF
\end{minipage} &
\begin{minipage}{0.130\linewidth}\centering
\includegraphics[width=\linewidth]{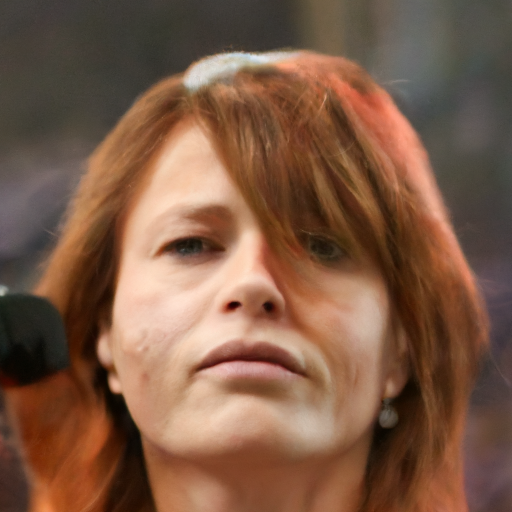}\
DifFace
\end{minipage} &
\begin{minipage}{0.130\linewidth}\centering
\includegraphics[width=\linewidth]{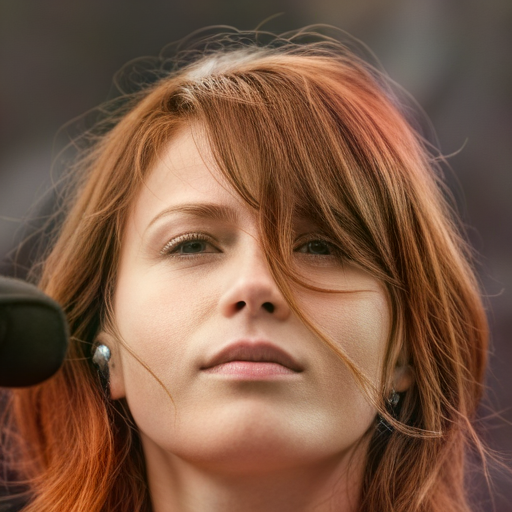}\
Authface
\end{minipage} &
\begin{minipage}{0.130\linewidth}\centering
\includegraphics[width=\linewidth]{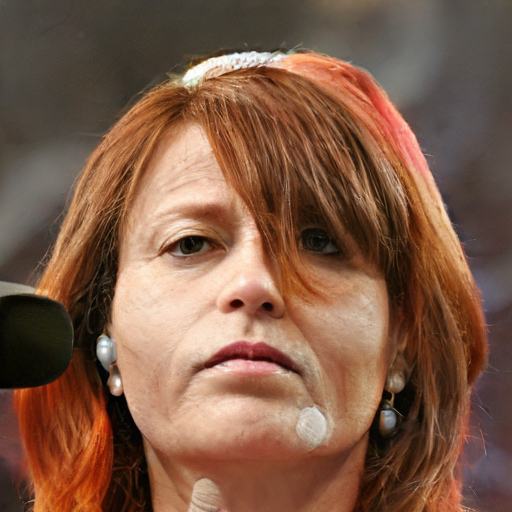}\
OSDFace
\end{minipage} &
\begin{minipage}{0.130\linewidth}\centering
\includegraphics[width=\linewidth]{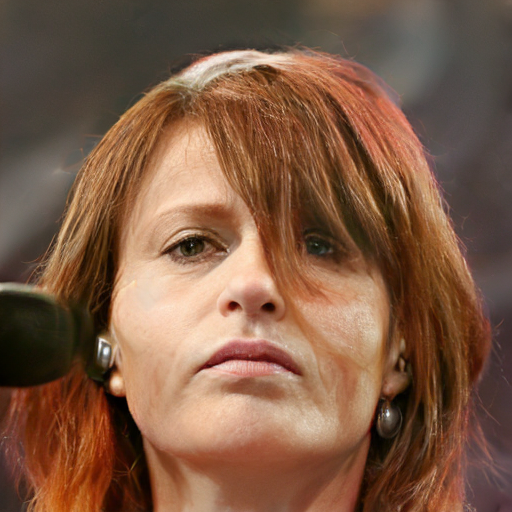}\
Ours
\end{minipage}
\end{tabular}
\caption{More Qualitative comparison on CelebA-Test dataset. From left to right: HQ, results of other methods, where CodeF stands for CodeFormer. Please zoom in for a better view.}
\label{fig:celebA_compare}
\end{figure*}

\noindent\textbf{Qualitative comparison on the CelebA-Test.} 
We provide additional qualitative comparisons on the CelebA dataset. As illustrated in Fig.~\ref{fig:celebA_compare}, in the first row, our model demonstrates higher fidelity in restoring eye details compared to other methods. In the second row, the skin textures and wrinkles generated by our approach appear more natural and are highly consistent with the ground truth. In the third row, only our method successfully reconstructs the earrings on both sides of the face. Furthermore, in the fourth and fifth rows, our method significantly outperforms the competing approaches in recovering fine hair details. Such superior performance in restoring fine facial features and local details fully validates the rationality and effectiveness of our proposed design.

\begin{figure*}[t]
\centering
\setlength{\tabcolsep}{2pt} % 水平间距
\renewcommand{\arraystretch}{0.0} % 垂直紧凑
\footnotesize % 标注字号

% 行1
\begin{tabular}{cccccccc}
\begin{minipage}{0.130\linewidth}\centering
\includegraphics[width=\linewidth]{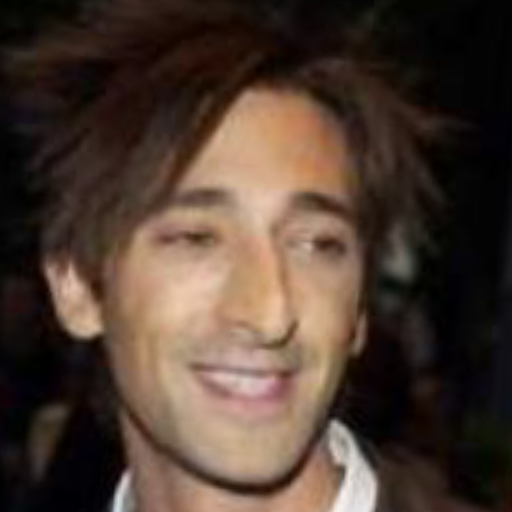}\
LQ
\end{minipage} &
% \begin{minipage}{0.124\linewidth}\centering
% \includegraphics[width=\linewidth]{images/00000040_LR.png}\
% LQ
% \end{minipage} &
\begin{minipage}{0.130\linewidth}\centering
\includegraphics[width=\linewidth]{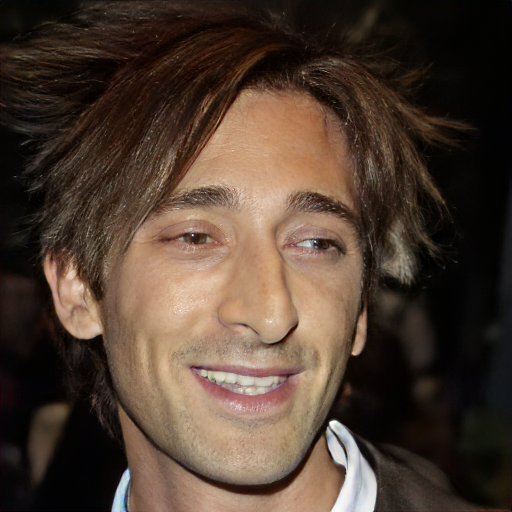}\
VQFR
\end{minipage} &
\begin{minipage}{0.130\linewidth}\centering
\includegraphics[width=\linewidth]{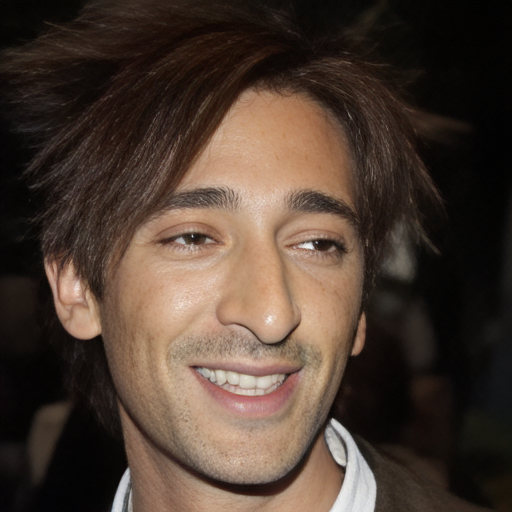}\
CodeF
\end{minipage} &
\begin{minipage}{0.130\linewidth}\centering
\includegraphics[width=\linewidth]{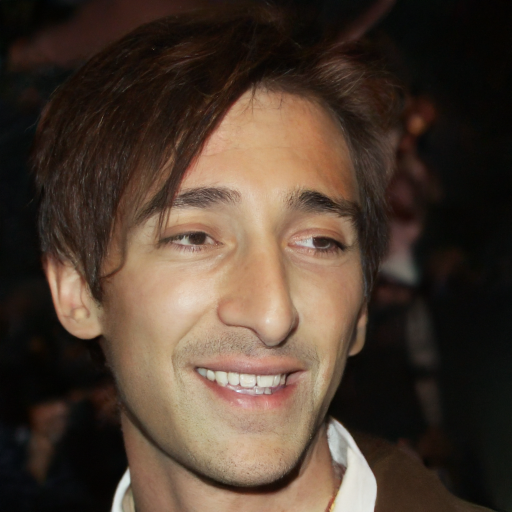}\
DifFace
\end{minipage} &
\begin{minipage}{0.130\linewidth}\centering
\includegraphics[width=\linewidth]{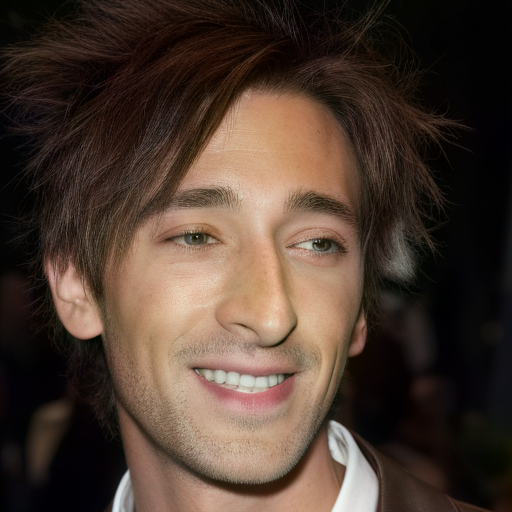}\
Authface
\end{minipage} &
\begin{minipage}{0.130\linewidth}\centering
\includegraphics[width=\linewidth]{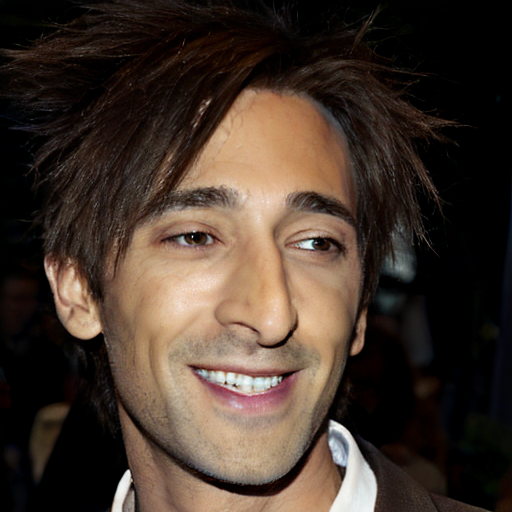}\
OSDFace
\end{minipage} &
\begin{minipage}{0.130\linewidth}\centering
\includegraphics[width=\linewidth]{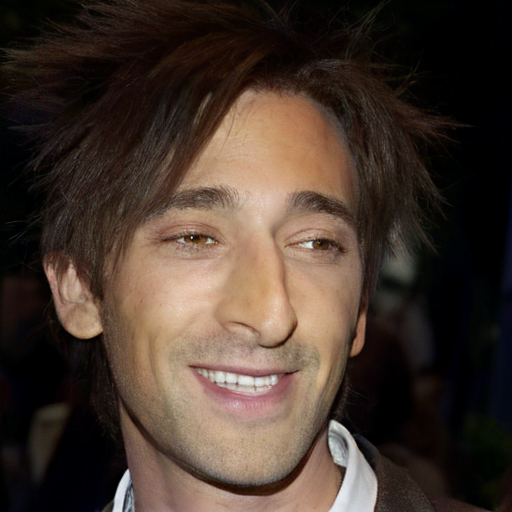}\
Ours
\end{minipage}
\end{tabular}

\vspace{2pt}

% 行2
\begin{tabular}{ccccccc}
\begin{minipage}{0.130\linewidth}\centering
\includegraphics[width=\linewidth]{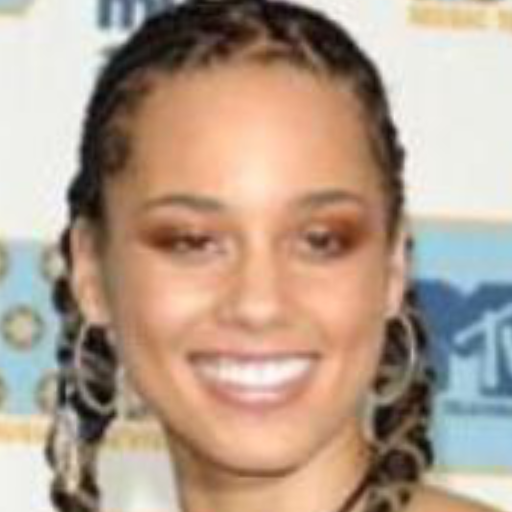}\
LQ
\end{minipage} &
% \begin{minipage}{0.124\linewidth}\centering
% \includegraphics[width=\linewidth]{images/00000044_LR.png}\
% LQ
% \end{minipage} &
\begin{minipage}{0.130\linewidth}\centering
\includegraphics[width=\linewidth]{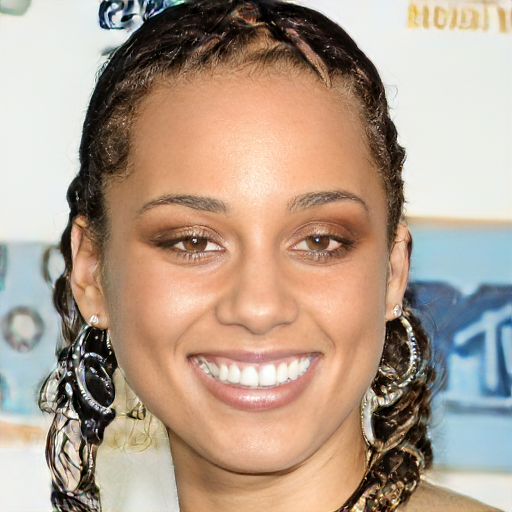}\
VQFR~
\end{minipage} &
\begin{minipage}{0.130\linewidth}\centering
\includegraphics[width=\linewidth]{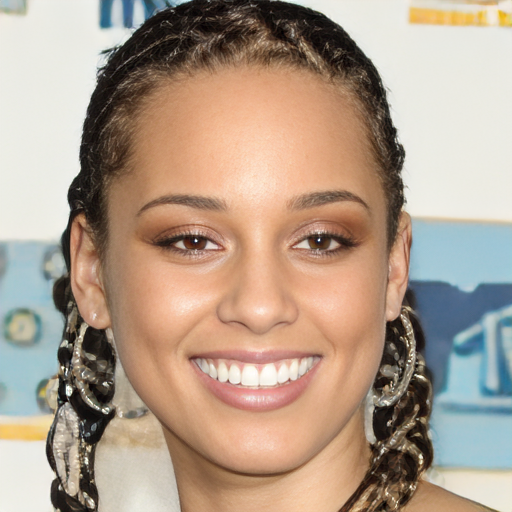}\
CodeF
\end{minipage} &
\begin{minipage}{0.130\linewidth}\centering
\includegraphics[width=\linewidth]{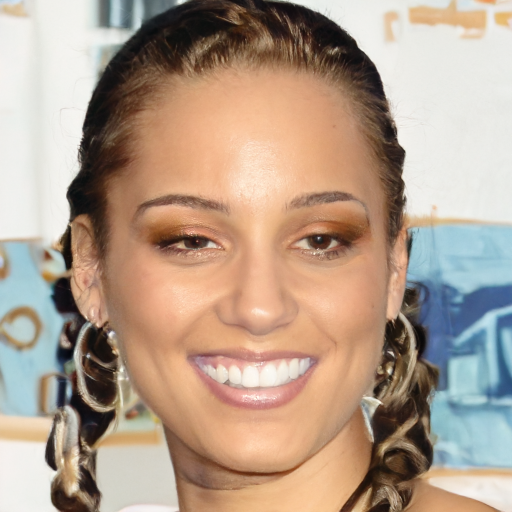}\
DifFace
\end{minipage} &
\begin{minipage}{0.130\linewidth}\centering
\includegraphics[width=\linewidth]{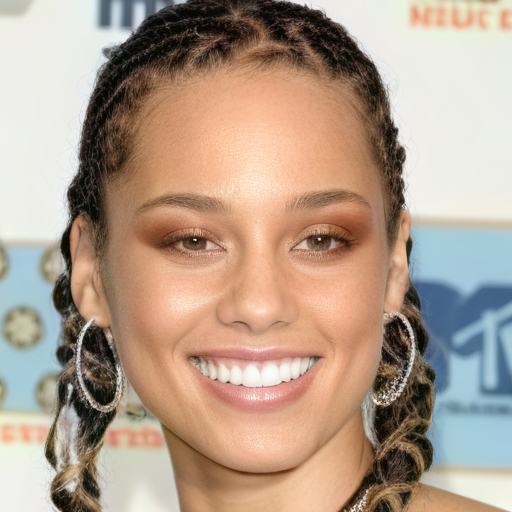}\
Authface
\end{minipage} &
\begin{minipage}{0.130\linewidth}\centering
\includegraphics[width=\linewidth]{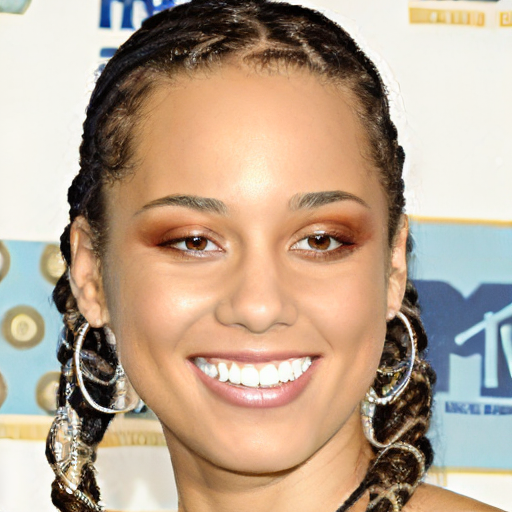}\
OSDFace
\end{minipage} &
\begin{minipage}{0.130\linewidth}\centering
\includegraphics[width=\linewidth]{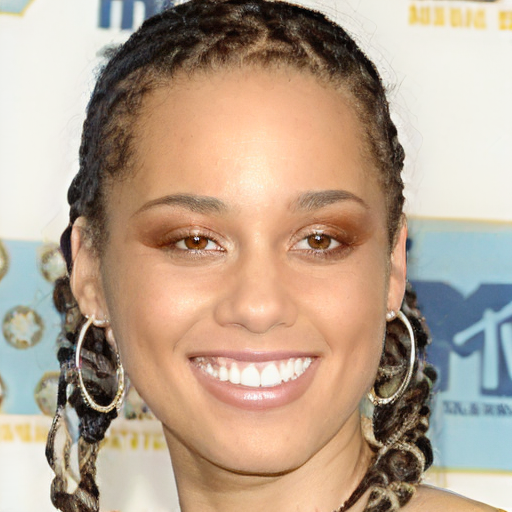}\
Ours
\end{minipage}
\end{tabular}

\vspace{2pt}

% 行3
\begin{tabular}{ccccccc}
\begin{minipage}{0.130\linewidth}\centering
\includegraphics[width=\linewidth]{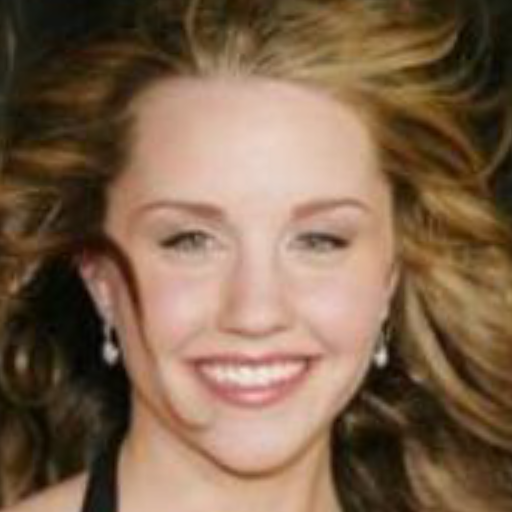}\
LQ
\end{minipage} &
% \begin{minipage}{0.124\linewidth}\centering
% \includegraphics[width=\linewidth]{images/00000057_LR.png}\
% LQ
% \end{minipage} &
\begin{minipage}{0.130\linewidth}\centering
\includegraphics[width=\linewidth]{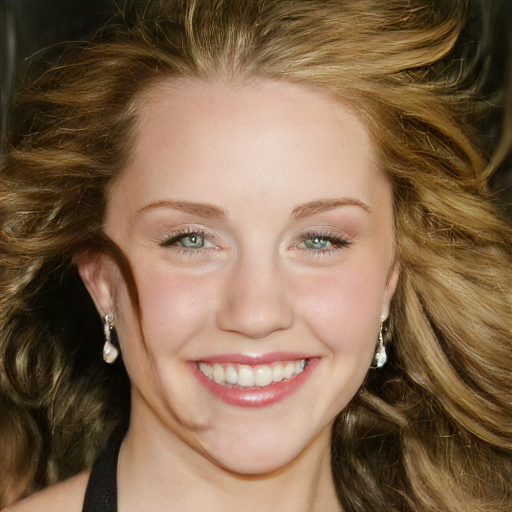}\
VQFR
\end{minipage} &
\begin{minipage}{0.130\linewidth}\centering
\includegraphics[width=\linewidth]{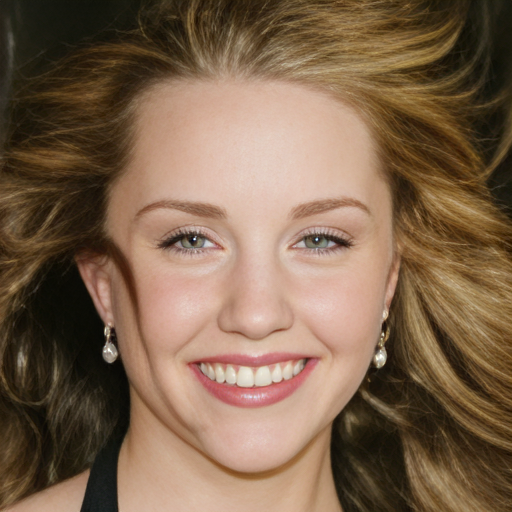}\
CodeF
\end{minipage} &
\begin{minipage}{0.130\linewidth}\centering
\includegraphics[width=\linewidth]{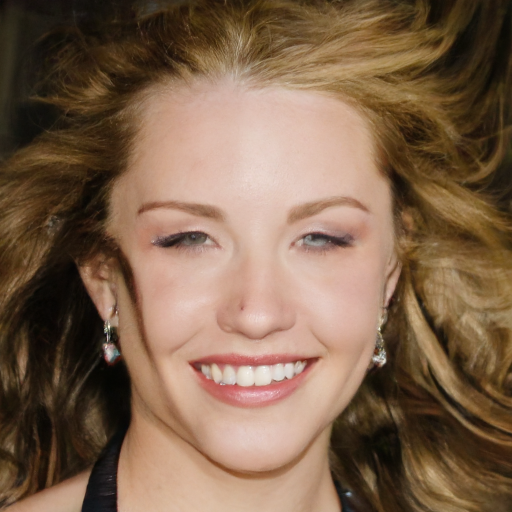}\
DifFace
\end{minipage} &
\begin{minipage}{0.130\linewidth}\centering
\includegraphics[width=\linewidth]{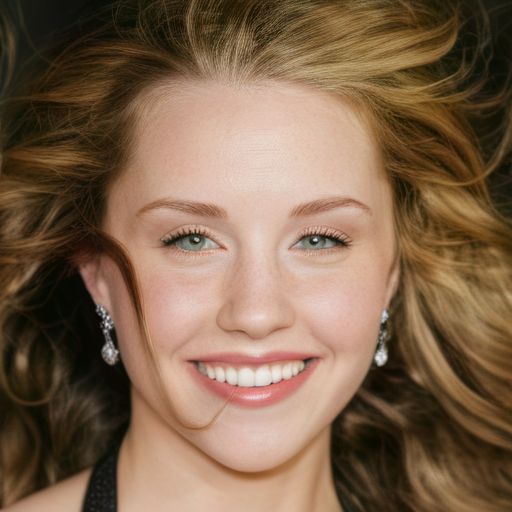}\
Authface
\end{minipage} &
\begin{minipage}{0.130\linewidth}\centering
\includegraphics[width=\linewidth]{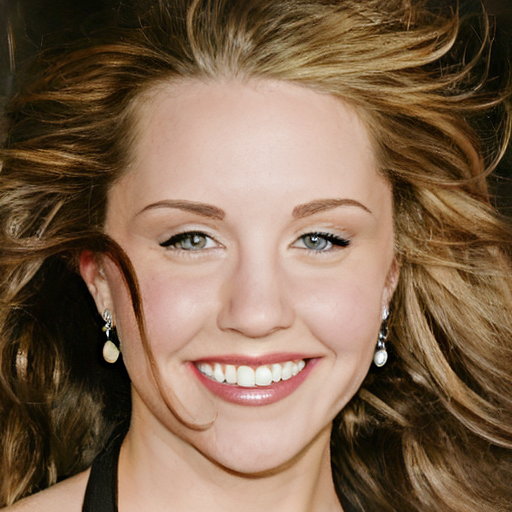}\
OSDFace
\end{minipage} &
\begin{minipage}{0.130\linewidth}\centering
\includegraphics[width=\linewidth]{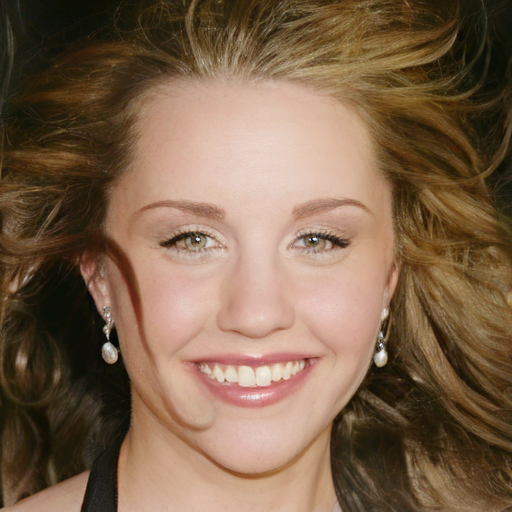}\
Ours
\end{minipage}
\end{tabular}
\caption{More Qualitative comparison on LFW-Test dataset. From left to right: LQ, results of other methods, where CodeF stands for CodeFormer. Please zoom in for a better view.}
\label{fig:LFW_compare}
\end{figure*}

\noindent\textbf{Qualitative comparison on the LFW-Test.} 
We conducted qualitative comparison of our method with other approaches on the real-world LFW dataset. As illustrated in Fig.~\ref{fig:LFW_compare}, specifically, in the first-row example, our method achieves more natural visual effects in restoring facial details such as teeth compared to OSDFace. In the second row, our model yields a more realistic reconstruction of the earrings. Furthermore, the third row demonstrates the superior overall restoration performance of our method, with the generated images closely resembling the original images in terms of color fidelity and global details.

\begin{figure*}[t]
\centering
\setlength{\tabcolsep}{2pt} % 水平间距
\renewcommand{\arraystretch}{0.0} % 垂直紧凑
\footnotesize % 标注字号

% 行1
\begin{tabular}{cccccccc}
\begin{minipage}{0.130\linewidth}\centering
\includegraphics[width=\linewidth]{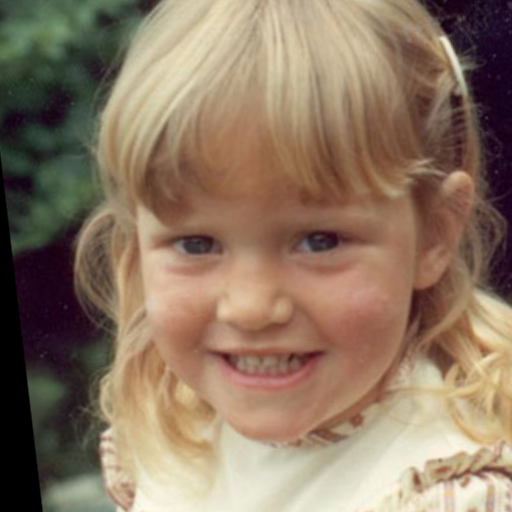}\
LQ
\end{minipage} &
% \begin{minipage}{0.124\linewidth}\centering
% \includegraphics[width=\linewidth]{images/00000040_LR.png}\
% LQ
% \end{minipage} &
\begin{minipage}{0.130\linewidth}\centering
\includegraphics[width=\linewidth]{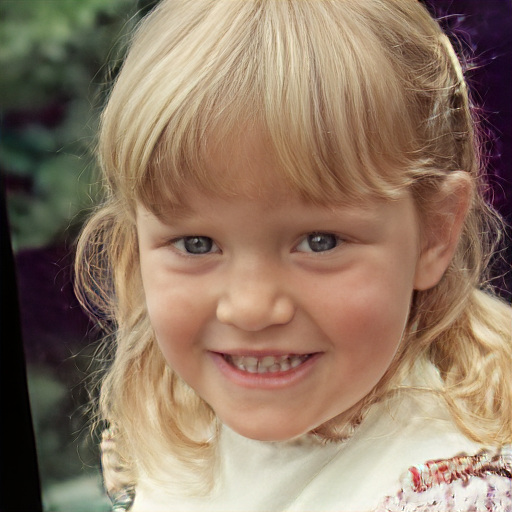}\
VQFR
\end{minipage} &
\begin{minipage}{0.130\linewidth}\centering
\includegraphics[width=\linewidth]{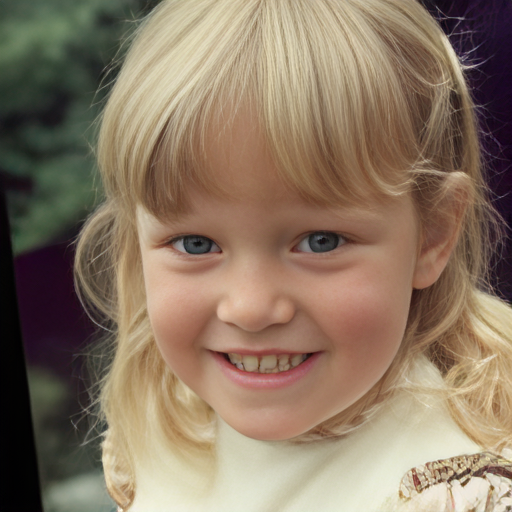}\
CodeF
\end{minipage} &
\begin{minipage}{0.130\linewidth}\centering
\includegraphics[width=\linewidth]{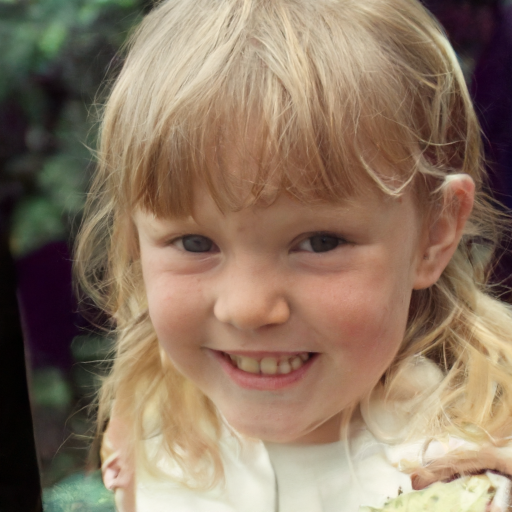}\
DifFace
\end{minipage} &
\begin{minipage}{0.130\linewidth}\centering
\includegraphics[width=\linewidth]{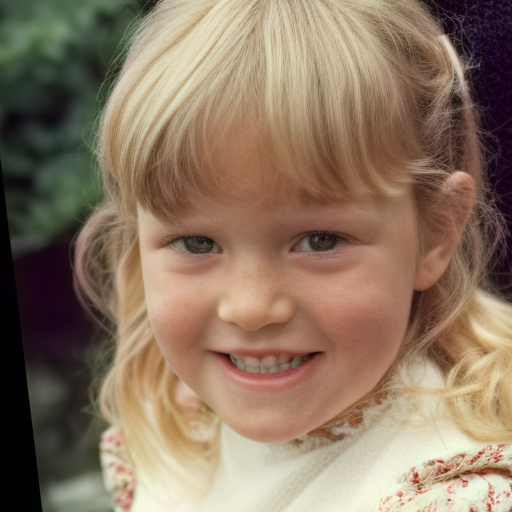}\
Authface
\end{minipage} &
\begin{minipage}{0.130\linewidth}\centering
\includegraphics[width=\linewidth]{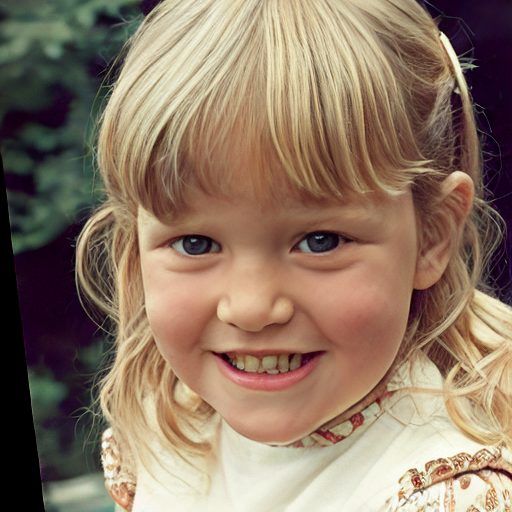}\
OSDFace
\end{minipage} &
\begin{minipage}{0.130\linewidth}\centering
\includegraphics[width=\linewidth]{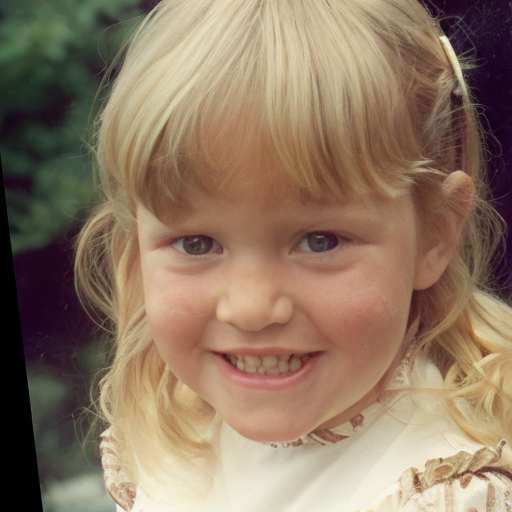}\
Ours
\end{minipage}
\end{tabular}

\vspace{2pt}

% 行2
\begin{tabular}{ccccccc}
\begin{minipage}{0.130\linewidth}\centering
\includegraphics[width=\linewidth]{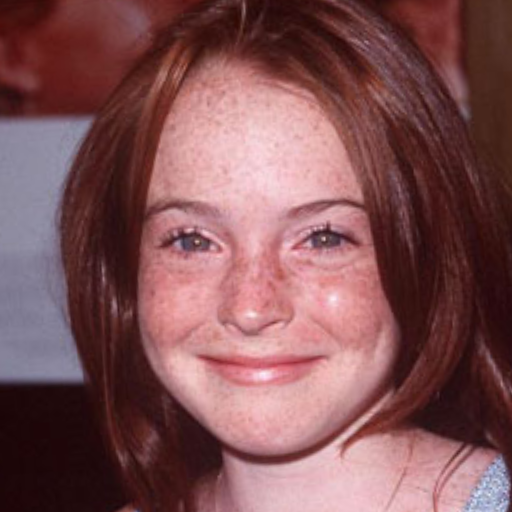}\
LQ
\end{minipage} &
% \begin{minipage}{0.124\linewidth}\centering
% \includegraphics[width=\linewidth]{images/00000044_LR.png}\
% LQ
% \end{minipage} &
\begin{minipage}{0.130\linewidth}\centering
\includegraphics[width=\linewidth]{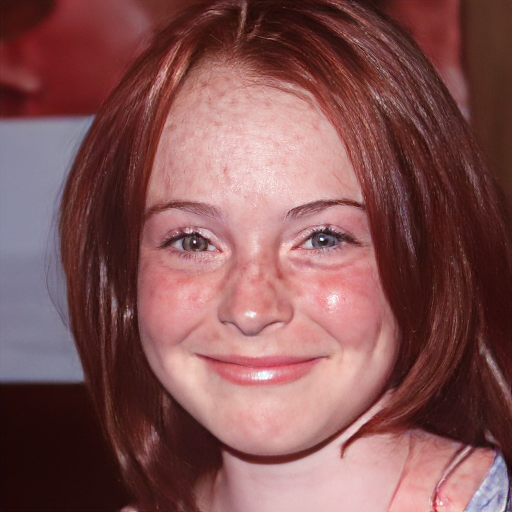}\
VQFR
\end{minipage} &
\begin{minipage}{0.130\linewidth}\centering
\includegraphics[width=\linewidth]{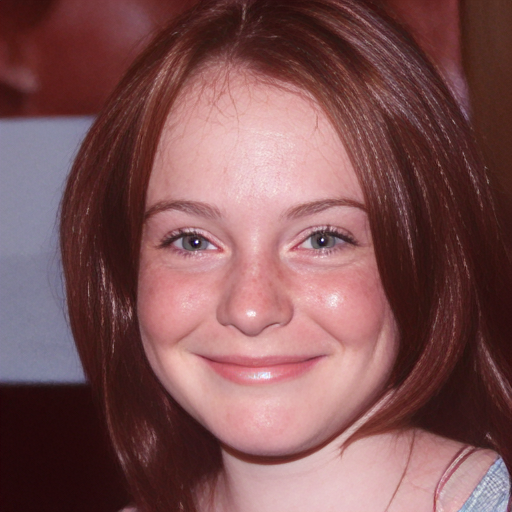}\
CodeF
\end{minipage} &
\begin{minipage}{0.130\linewidth}\centering
\includegraphics[width=\linewidth]{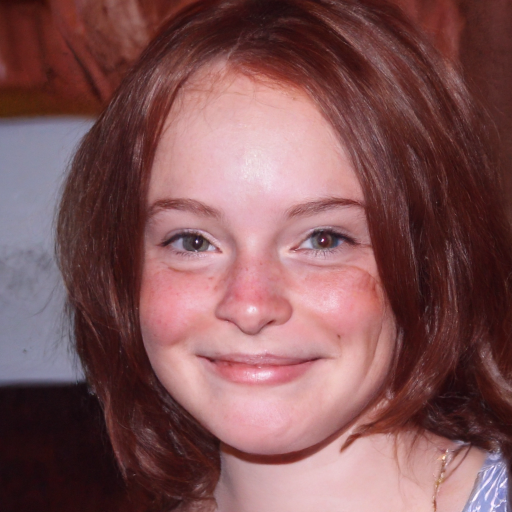}\
DifFace
\end{minipage} &
\begin{minipage}{0.130\linewidth}\centering
\includegraphics[width=\linewidth]{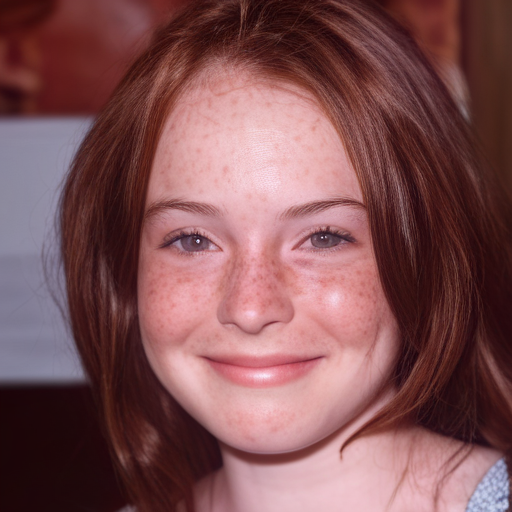}\
Authface
\end{minipage} &
\begin{minipage}{0.130\linewidth}\centering
\includegraphics[width=\linewidth]{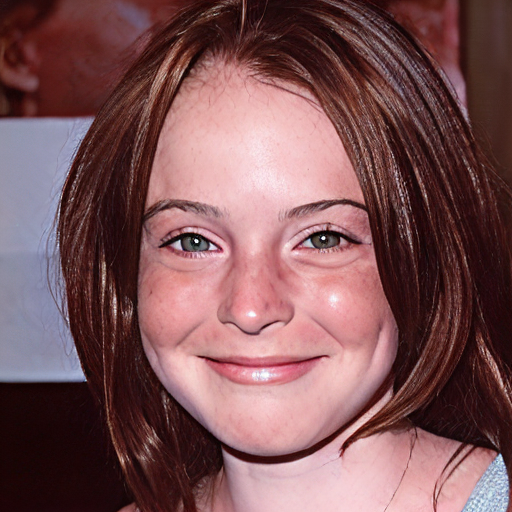}\
OSDFace
\end{minipage} &
\begin{minipage}{0.130\linewidth}\centering
\includegraphics[width=\linewidth]{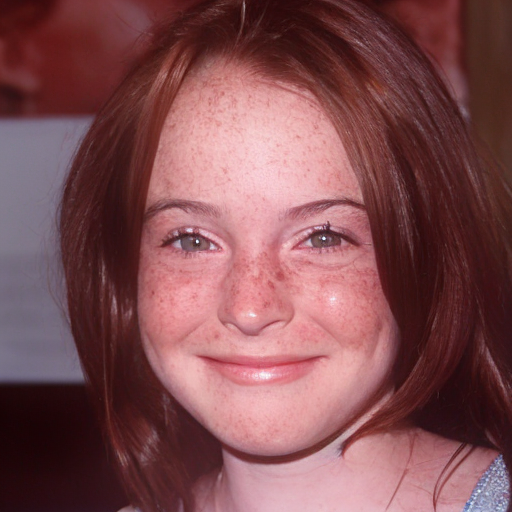}\
Ours
\end{minipage}
\end{tabular}

\vspace{2pt}

% 行3
\begin{tabular}{ccccccc}
\begin{minipage}{0.130\linewidth}\centering
\includegraphics[width=\linewidth]{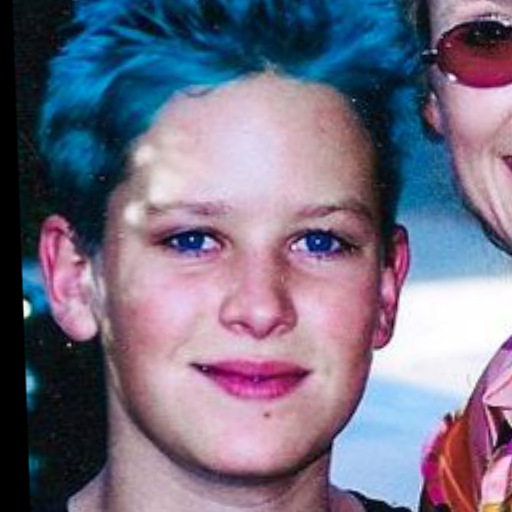}\
LQ
\end{minipage} &
% \begin{minipage}{0.124\linewidth}\centering
% \includegraphics[width=\linewidth]{images/00000057_LR.png}\
% LQ
% \end{minipage} &
\begin{minipage}{0.130\linewidth}\centering
\includegraphics[width=\linewidth]{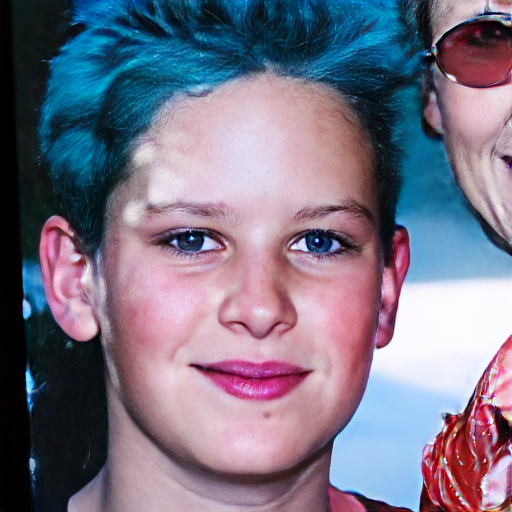}\
VQFR
\end{minipage} &
\begin{minipage}{0.130\linewidth}\centering
\includegraphics[width=\linewidth]{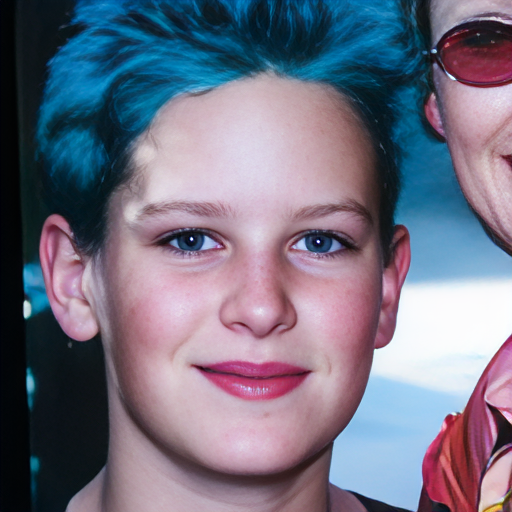}\
CodeF
\end{minipage} &
\begin{minipage}{0.130\linewidth}\centering
\includegraphics[width=\linewidth]{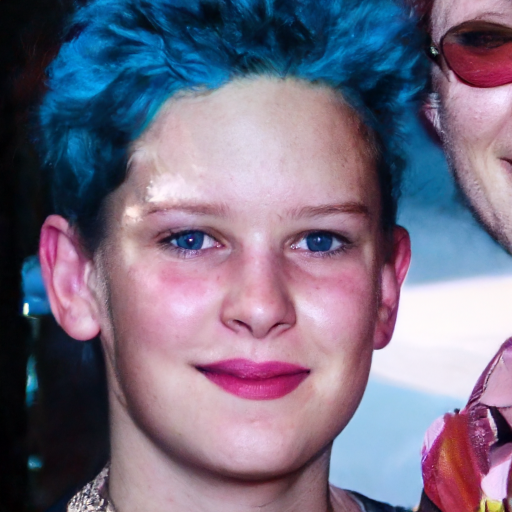}\
DifFace
\end{minipage} &
\begin{minipage}{0.130\linewidth}\centering
\includegraphics[width=\linewidth]{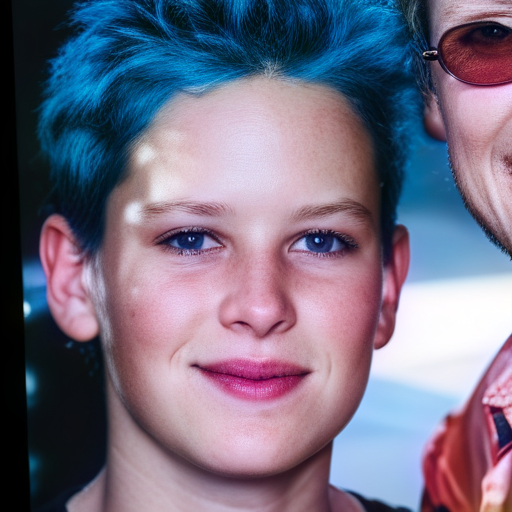}\
Authface
\end{minipage} &
\begin{minipage}{0.130\linewidth}\centering
\includegraphics[width=\linewidth]{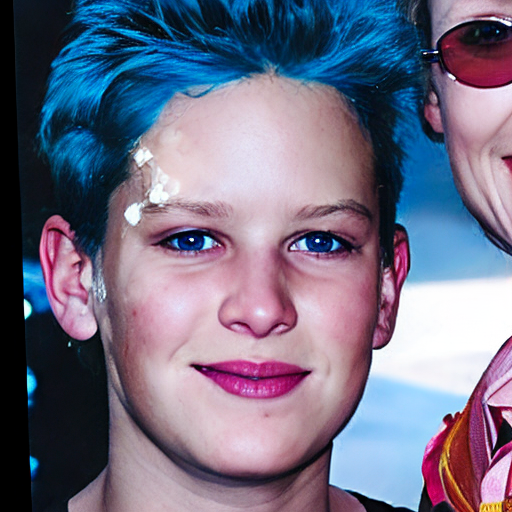}\
OSDFace
\end{minipage} &
\begin{minipage}{0.130\linewidth}\centering
\includegraphics[width=\linewidth]{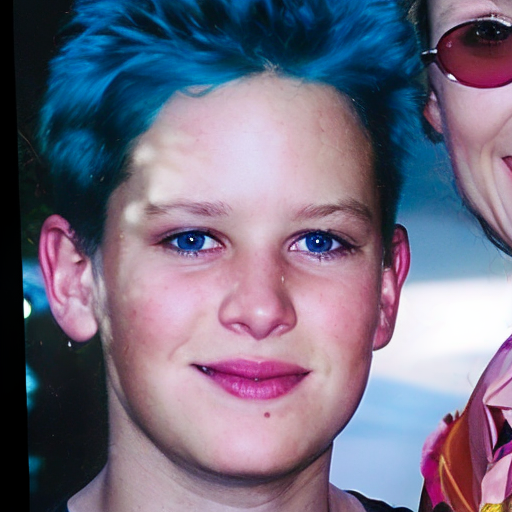}\
Ours
\end{minipage}
\end{tabular}
\caption{More Qualitative comparison on CeleChild-Test dataset. From left to right: LQ, results of other methods, where CodeF stands for CodeFormer. Please zoom in for a better view.}
\label{fig:Child_compare}
\end{figure*}

\noindent\textbf{Qualitative comparison on the CeleChild-Test.} 
We conducted qualitative comparisons with other methods on the real-world CeleChild dataset. As illustrated in Fig.~\ref{fig:Child_compare}, specifically, in the first-row example, compared to other methods, our approach achieves more natural visual effects in recovering hair details and is closer to the ground truth. In the second row, only our method naturally and completely preserves the freckles on the girl's face, whereas other methods either treat them as noise, resulting in over-smoothing, or generate unnatural textures. In the third row, OSDFace fails to properly handle the natural illumination on the left side of the boy's face, leading to hallucinated outputs, whereas our method accurately and reasonably restores this lighting distribution.

\begin{table}[htbp]
  \centering
  \begin{tabular}{l|ccccc}
    \toprule
    Methods & PGDiff~\cite{yang2023pgdiff} & DifFace~\cite{yue2024difface} & AuthFace~\cite{liang2025authface} & OSDFace~\cite{wang2025osdface} & Ours\\
    \midrule
    Step & 1,000 & 250 & 50 & 1 & 1 \\
    Time (s) & 73.01 & 4.120 & 4.205 & 0.488  & 0.988\\
    Param (M) & 176.4 & 175.4 & 4919.9 & 978.4 & 31413\\
    MACs (G) & 480,997 & 18,682 & 113667 & 2,132 & 18224 \\
    \bottomrule
  \end{tabular}
  \vspace{2mm} % 调整标题与表格的间距
  \caption{Complexity comparison during inference. All models are tested with an input image size of $512 \times 512$.}
  \label{tab:complexity}
\end{table}

\subsection{The computational cost}
\label{appendix:a.3}
Tab.~\ref{tab:complexity} presents a comparison of the computational costs between our method and other generative model-based approaches. Benefiting from the one-step inference mechanism, our method significantly outperforms traditional multi-step inference methods in terms of inference latency. Compared to OSDFace, a counterpart one-step inference method, our approach experiences a slight increase in inference time due to the incorporation of large-scale parameter models such as Qwen-Image. Nevertheless, the overall computational cost remains within a reasonable range, without incurring an order-of-magnitude increase in time overhead.

% \begin{table*}[t]
% \centering
% \small
% \caption{Matched-components and matched-compute comparison. This table separates the effect of the proposed representation-fusion design from the effect of using additional priors or larger computation.}
% \label{tab:matched_compute}
% \setlength{\tabcolsep}{3pt}
% \renewcommand{\arraystretch}{1.1}
% \begin{tabular}{l|ccc|cccc}
% \toprule
% \multirow{2}{*}{Method}
% & \multirow{2}{*}{Inter. restorer}
% & \multirow{2}{*}{DINO cond.}
% & \multirow{2}{*}{Fusion}
% & \multicolumn{4}{c}{CelebA-Test} \\
% \cmidrule(lr){5-8}
% & & & & LPIPS$\downarrow$ & DISTS$\downarrow$ & TOPIQ$\uparrow$ & Deg.$\downarrow$ \\
% \midrule
% OSDFace + LR images& \xmark & \xmark & --& -- & -- & -- & -- \\
% OSDFace + SR images & \cmark & \xmark & -- & -- & -- & -- & -- \\
% OSDFace + DINO condition & \cmark & \cmark & concat & -- & -- & -- & -- \\
% HDRFace-small & \cmark & DINOv3-S/B & SDFM & -- & -- & -- & -- \\
% HDRFace-full & \cmark & DINOv3-L & SDFM & -- & -- & -- & -- \\
% \bottomrule
% \end{tabular}
% \end{table*}

\subsection{Ablation study on our guidance model choice}
\begin{table*}[t]
\small
\centering
\renewcommand{\arraystretch}{1.1}
\setlength{\tabcolsep}{1pt}
\vspace{-0.15in}
\caption{Comparison of metrics with different guidance models, where C-IQA stands for CLIP-IQA.}
\begin{tabular}{l|cc|cc|ccc}
\Xhline{1.1pt}
\multirow{2}{*}{Methods}
& \multicolumn{7}{c}{\textbf{CelebA-Test}} \\
\cline{2-8}
 & Deg. $\downarrow$ & LMD $\downarrow$ & SSIM $\uparrow$ & PSNR $\uparrow$ & LPIPS $\downarrow$ & DISTS$\downarrow$ & TOPIQ$\uparrow$ \\
\hline 
OSDFace & 35.71 & 2.223 & 0.6351 & 22.12 & 0.2561 & 0.1660 & 0.4601\\
% \rowcolor[gray]{0.9}  
Ours+CodeFormer   & 31.03 & 1.971 & 0.6678 & 23.03 & 0.2311 & 0.1489 & 0.5125 \\
% \rowcolor[gray]{0.9} 
Ours+Qwen-Image  & \textbf{30.12} & \textbf{1.939} & \textbf{0.6782} & \textbf{23.62} & \textbf{0.2224} & \textbf{0.1433} & \textbf{0.5344} \\
\Xhline{1.1pt}
\end{tabular}
\label{Table: ablation method}
\end{table*}
Given that the guidance model inevitably introduces additional computational overhead, and non-diffusion models such as CodeFormer~\cite{zhou2022towards} are computationally lighter than Qwen-Image, we conduct an ablation study to investigate whether employing a lightweight model can achieve comparable performance. As shown in Tab.~\ref{Table: ablation method}, the results indicate that although utilizing CodeFormer yields certain performance improvements over the baseline, its overall efficacy still falls short of our proposed method. This further demonstrates the necessity and rationality of incorporating Qwen-Image into our framework.

\subsection{Failure Case Analysis}
\label{appendix:failure_analysis}
We further analyze representative failure cases to clarify the practical boundary of HDRFace under extremely severe degradations. The purpose of this analysis is not to suggest that our method fails frequently, but to make explicit where the ill-posed nature of face restoration still dominates. As demonstrated by the quantitative results and the extensive qualitative comparisons above, HDRFace consistently improves restoration quality and identity preservation across different settings. The cases in this section instead correspond to hard examples where the LQ input has already removed identity-critical cues, making the exact HR target ambiguous even for a strong generative prior.

\begin{figure*}[t]
\centering
\includegraphics[width=0.9\linewidth]{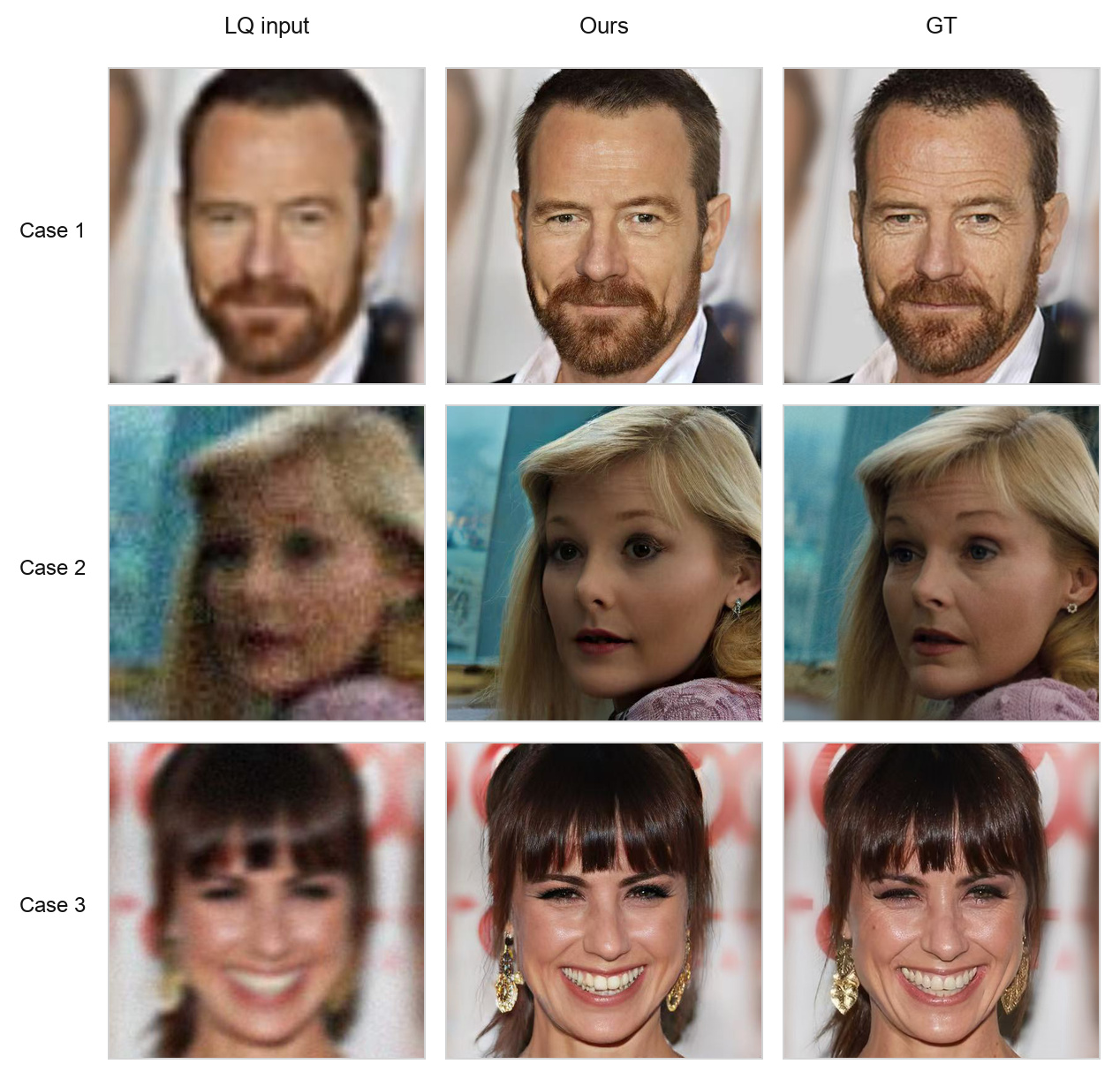}
\caption{Representative failure cases of HDRFace under severe degradations. Our method still restores clear and realistic facial structures, but local identity-sensitive details may deviate from the ground truth when the LQ input loses critical information.}
\label{fig:failure_cases}
\vspace{-0.8em}
\end{figure*}

As shown in Fig.~\ref{fig:failure_cases}, HDRFace remains able to generate visually plausible faces even when the LQ observations are highly blurred or noisy. In Case 1 and Case 2, the inputs preserve coarse pose, face layout, and color context, but obscure fine-grained cues such as eye contours, skin wrinkles, hair boundaries, beard texture, and accessories. Our restored results recover coherent facial structures and realistic high-frequency details, yet still exhibit residual deviations from the GT image, including over-smoothed skin, slightly shifted eye shapes, and imperfect reconstruction of beard, hair, or earrings. In Case 3, the global identity and expression are largely preserved, but very local structures such as bangs, eye corners, teeth alignment, and earrings can still be hallucinated or over-sharpened.

These examples reveal a bounded failure mode rather than a fundamental weakness of the proposed framework. HDRFace uses high-dimensional representations to provide stronger facial priors than the LQ input alone, but no representation can uniquely recover details that are completely absent from the observation. Moreover, because the current pipeline relies on a single intermediate restoration, incorrect or over-confident details in this intermediate result may occasionally be propagated into the final output. SDFM mitigates this issue by adaptively balancing LR-side structural cues and intermediate-result-side detail cues, but it does not explicitly estimate uncertainty for highly ambiguous regions. A natural extension is to generate multiple intermediate candidates and select or fuse them according to identity consistency, local reliability, or uncertainty estimates, which could further reduce these rare but informative failure cases.

% \clearpage
% \newpage
% \input{checklist.tex}

\end{document}